\documentclass{article}

\usepackage{microtype}
\usepackage{graphicx}
\usepackage{subcaption}
\usepackage{booktabs} %
\usepackage{hyperref}
\usepackage{url}
\usepackage[table]{xcolor}
\usepackage{multirow}
\usepackage{caption} %
\usepackage{comment} 
\usepackage{wrapfig}

\usepackage{hyperref}

\usepackage[main, final]{neurips_2026}

\usepackage[utf8]{inputenc} %
\usepackage[T1]{fontenc}    %
\usepackage{hyperref}       %
\usepackage{url}            %
\usepackage{booktabs}       %
\usepackage{amsfonts}       %
\usepackage{nicefrac}       %
\usepackage{microtype}      %
\usepackage{xcolor}         %
\usepackage{amsmath}
\usepackage{amssymb}
\usepackage{mathtools}
\usepackage{amsthm}

\usepackage[capitalize,noabbrev]{cleveref}
\title{Test-Time Defense Against Adversarial Attacks \\via Stochastic Resonance of Latent Ensembles}

\author{%
  Dong Lao\thanks{Project page: \url{https://donglao.github.io/sr/}. Correspondence: \href{mailto:dong.lao@lsu.edu}{dong.lao@lsu.edu}.} \\
  LSU
  \And
  Yuxiang Zhang \\
  UCLA
  \And
  Haniyeh Ehsani Oskouie\\
  UCLA
  \And
  Yangchao Wu\\
  AWS 
  \AND
  Alex Wong \\
  Yale
  \And
  Stefano Soatto \\
  AWS \& UCLA
}

\begin{document}

\maketitle

\begin{abstract}
We propose a test-time defense mechanism against adversarial attacks: Unlike existing methods that rely on feature filtering or smoothing, which can lead to information loss, we propose to ``combat noise with noise'' by leveraging stochastic resonance to enhance robustness while minimizing information loss. Our approach introduces small translational perturbations to the input image, aligns the transformed feature embeddings, and aggregates them before mapping back to the original reference image. This can be expressed in a closed-form formula, which can be deployed on diverse existing network architectures without introducing additional network modules or fine-tuning for specific attack types. The resulting method is entirely training-free, architecture-agnostic, and attack-agnostic. Empirically, the method achieves state-of-the-art robustness on image classification and provides the first generic test-time defense for dense prediction tasks, including stereo matching, optical flow, and monocular depth estimation. Across adversarial attacks, it recovers up to 68.1\% of the accuracy loss on image classification, 71.9\% on stereo matching (MAE), 29.2\% on optical flow (EPE), 68.3\% on depth estimation (SqRel),  and 83.8\% on vision-language alignment (cosine similarity).

\end{abstract}
\section{Introduction}
\label{sec:introduction}

Most deep neural networks in use today are deterministic maps from a fixed-size input to a fixed-size feature vector. In auto-regressive Transformer models, that vector encodes the next element (token) in the input sequence. Similarly, in convolutional architectures, that vector may encode the input data. In either case, the output vector is often highly sensitive to perturbations of the input, and one can intentionally choose these imperceptible perturbations {\em adversarially} so as to maximize the change in the output \cite{goodfellow2014explaining}. In some cases, the same perturbation can even be disruptive for a large number of possible inputs \cite{moosavi2017universal}, exploiting the convoluted geometry of the decision boundary imposed by such trained models \cite{tramer2017space}. This spurious sensitivity could be exploited adversarially to disrupt the operation of a model. 

From a classical perspective of signal processing, adversarial perturbations of images appear as small high-frequency ``noise'' resembling {\em aliasing} artifacts. These are imperceptible since the human visual system easily discounts them on account of their poor fit to the ``ecological statistics'' of natural images \cite{gibson1978ecological}.
Classical sampling theory prescribes anti-aliasing by low-pass filtering the data, removing information along with the artifacts. Low-pass filtering consists of spatial averaging of the perturbed data with respect to a chosen kernel, typically a Gaussian or a constant (``pillbox''). Alternatively, one could ``denoise'' the embeddings by averaging output vectors, also a lossy operation. The choice of the kernel should match the statistics of the perturbations, which sets up a cat-and-mouse game where the adversary can easily modify the perturbations to bypass the anti-aliasing filter, namely {\em adaptive} attacks \cite{tramer2020adaptive}, and the model needs to constantly be updated to ``anti-alias'' new forms of adversarial perturbations. In the context of deep neural networks, existing defense methods that manipulate feature representations \cite{xie2019feature,bai2021improving,yan2021cifs,kim2023feature} fundamentally adhere to this paradigm. Thus, despite substantial engineering efforts, these methods remain inherently vulnerable to adaptive attacks because they rely on pre-defined filtering strategies that are fixed at inference time.

\textbf{Desiderata:} 
To break this cycle, we advocate a method to mitigate the effect of adversarial perturbation that (i) operates at test-time, without the need to update the model weights, and (ii) does not entail information loss associated with direct or indirect spatial filtering. In addition, it would be ideal if this method could (iii) be applied to existing network architectures without modifications, and (iv) be agnostic to the specific type of adversarial perturbation.

\textbf{Stochastic Resonance} is a technique that resolves a quantized signal below the quantization level, by quantizing and ensembling perturbed versions of the signal \cite{benzi1981mechanism}. It has been used extensively in cochlear implants, where the resolution of the digital circuitry is limited\cite{stocks2002application}. It has also been used to `super-resolve' Vision Transformer embeddings, where entire patches are encoded into a vector, which is computed at a coarsely subsampled grid \cite{lao2024sub}. In this paper, instead, we use Stochastic Resonance for the opposite purpose, {\em not} to super-resolve the quantized signal, but to perform latent ensembling to mitigate the effects of adversarial perturbations.

Rather than averaging the data, or averaging their embedding as in classical denoising, {\em we average transformed embeddings} in latent space. This averaging is performed over small transformations sampled at random or deterministically from the group of planar translations, by computing the encoding of the transformed image, and then mapping the encoding back through the push-forward of the inverse transformation. This process can be expressed as a single formula in \eqref{eq:SR}. Since the embedding is typically computed on a coarse grid, but the transformations are sampled on the native lattice of the image, the resulting embedding is free of spatial averaging artifacts. As with other uses of Stochastic Resonance, the effect is seemingly paradoxical as {\em we combat noise with noise}: We apply purposeful perturbations to eliminate the effect of adversarial perturbations.

Our method can be viewed as marginalizing the translation group in latent space with respect to a chosen prior. We choose the simplest, which is the constant prior. The purposeful perturbations alter the spatial sampling, and the ensembling in latent space averages out the effect of sampling artifacts, thwarting the effect of adversarial perturbations. It is as if we were given images with different `noise', except that the noise in question is not the adversarial perturbation, but the splinters of adversarial perturbations obtained by different spatially quantized versions of the perturbed image, due to the translational perturbations, which are then averaged out by the latent ensembling.

\textbf{Outcomes:} Our method fulfills the desiderata (i)-(iv) laid out earlier: (i) It does not require training or fine-tuning; (ii) it minimizes information loss by latent ensembling of perturbed embeddings; (iii) it can be applied to different network architectures and tasks, \emph{including} networks already equipped with different defense techniques like adversarial training, and (iv) is agnostic to the specific perturbation. We evaluate our method across six diverse tasks, including four dense prediction tasks. Stochastic resonance drastically improves adversarial robustness without training or architecture-specific modifications, recovering up to 68.1\% of the accuracy loss in classification, 71.9\% in stereo matching (MAE), 29.2\% in optical flow (EPE), 68.3\% in depth estimation (SqRel), and 83.8\% in vision-language alignment (cosine similarity). To our knowledge, this is the first test-time defense to show consistent gains across such a broad range of settings, especially on dense prediciton tasks.

\begin{figure*}[t]
\centering
\includegraphics[width=0.88\textwidth]{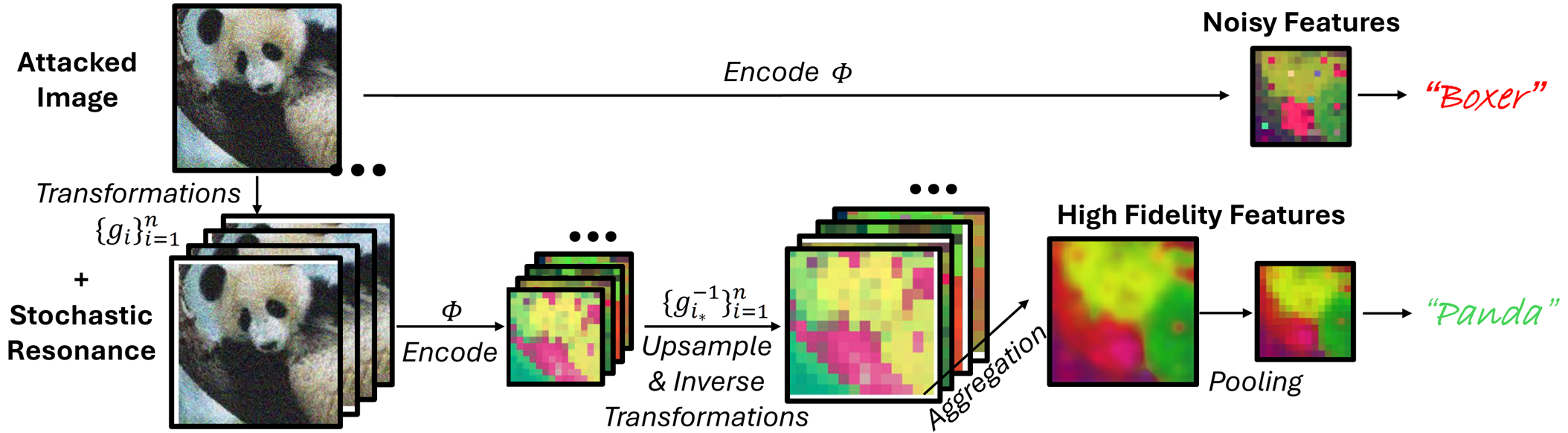}
    \caption{\small {\bf Diagram.} Neural networks are sensitive to perturbations in the input space, which adversarial attacks exploit to manipulate outputs. Existing defense focuses on filtering out unreliable features or denoising either the input or the features. Instead, we propose a novel defense based on stochastic resonance. Controlled transformations are introduced to the input. Features are then aggregated after inverting these transformations. The resulting test-time defense method requires no training and is compatible with diverse architectures. Notably, it not only improves robustness against attacks but also increases the difficulty of crafting adversarial examples, even when the attacker is aware of whether and how stochastic resonance is being used (adaptive adversary).
}
    \label{fig:schematic}
    
\end{figure*}

\section{Related Work}
\label{sec:related}

\noindent\textbf{Adversarial Training as Defense.}
Adversarial training increases the robustness of the model by training it with adversarially augmented images. The popular attack methods used are Fast Gradient Sign Method (FGSM) \cite{goodfellow2014explaining}, and Projected Gradient Descent (PGD) \cite{madry2017towards}. 
ALP \cite{kannan2018adversarial} Minimizes the difference between the logits of pairs of clean and adversarially augmented images. TRADES \cite{zhang2019theoretically} decomposes prediction error for adversarial examples into natural error and boundary error to improve adversarial robustness. MART \cite{wang2019improving} improves adversarial robustness by considering misclassified natural examples during training. Subsequent work \cite{cai2018curriculum, zhang2020attacks, wang2022self, jin2022enhancing, ghiasvand2024robust, xu2024autolora,gopalboosting} improves adversarial robustness with curriculum learning \cite{bengio2009curriculum}, model ensembling, critical parameter selection, and gradient tracking. 
On the other hand, some methods learn robust feature representation through a modified architecture or feature manipulation. \cite{galloway2019batch, benz2021batch, wang2022removing} investigate the effect of batch normalization on adversarial robustness. \cite{dhillon2018stochastic, madaan2020adversarial} prunes certain activations in the network that are susceptible to adversarial attacks.
\cite{xiao2019enhancing} keep k-features with the largest magnitude and deactivate everything else.
\cite{zoran2020towards} uses an attention mask to highlight robust regions on the feature.
Feature Denoising (FD) \cite{xie2019feature} uses classical denoising techniques to deactivate abnormal activations.
\cite{bai2021improving, yan2021cifs} proposed Channel Activation Suppression (CAS) and Channel-wise Importance-based Feature Selection (CIFS) to deactivate feature channels that are vulnerable to attacks. 
\cite{kim2023feature} improves the robustness with Feature Separation and Recalibration (FSR).
Our method also operates in feature space, but purely at test-time. While we choose some as baselines, our method can work in conjunction with any aforementioned methods.

\noindent\textbf{Adversarial Purification as Defense}
Another line of work focuses on purifying or augmenting the images before they are used as input. \cite{tang2024robust, yeh2024test, tsai2023test}  train an FGSM robust classifier, a diffusion model, or a mask auto-encoder, respectively, to purify adversarial examples.
\cite{wang2021fighting} optimizes both the model and the input to minimize the entropy of model predictions to adapt to changing attacks. 
\cite{cohen2024simple} trains a random forest predictor to ensemble outputs from test-time augmented images.
These works involve training a new model or updating the original model, while our method is purely test-time and does not require any training. 
\cite{perez2021enhancing} ensembles model output from different augmentations, which is a special case of our method, as the ensemble is performed solely on the output, while we can ensemble at any layer, which both saves computational cost and achieves higher performance. 
Notably, we recognize that above methods focus solely on classification as a task for adversarial attacks. Through extensive experiments, we demonstrate that our method can not only perform well on classification, but also on dense prediction tasks like optical flow, and stereo matching.

\noindent\textbf{Stochastic Resonance (SR)} was proposed by \cite{benzi1981mechanism} and first applied in climate dynamics \citep{benzi1982stochastic} and later in signal processing \citep{wellens2003stochastic, kosko2001robust, chen2007theory} and acoustics \citep{shu2016application, wang2014adaptive}. While several works \cite{zhang2022adversarially, ukita2023adversarial} introduce stochasticity within feature space, the most closely related prior effort is \cite{lao2024sub}, which applies stochastic resonance to “super-resolve’’ Vision Transformer (ViT) embeddings. In this work, we instead use SR to mitigate adversarial attacks. Since SR has been developed specifically to address {\em quantization artifacts}, it has never before been used to mitigate classes of perturbations beyond aliasing. Our novel use of the technique leverages the fact that group transformations and spatial quantization preserve the statistics of natural images, which are heavy-tailed, but do not preserve the statistics of adversarial perturbations. We are also closely related to  \cite{duan2026fisheye3r,gangopadhyay2025extending,gangopadhyay2026from,lao2024diffeomorphic,lao2024sub,wu2024augundo} in use of group transformations. 

\section{Method}
\label{sec:method}

\noindent\textbf{Formalization.}
An image $x:\Lambda\to\mathbb{R}^k$ is a sampled signal on a finite pixel lattice $\Lambda\subset\mathbb{Z}^2$, with $k$ color channels ($k=3$ for RGB). At the layer where we apply stochastic resonance, the encoder $\phi$ produces a spatial feature map
\[
\phi(x):\tilde{\Lambda}\to\mathbb{R}^K,
\]
where $\tilde{\Lambda}$ is the feature lattice and $K$ is the number of feature channels.

Let $\tau_{u,v}(p)=p+(u,v)$ denote a translation of spatial coordinates. We write $g_{u,v}$ for its action on an image:
\[
[g_{u,v}(x)](p)=x_{\mathrm{ext}}\bigl(\tau_{u,v}^{-1}(p)\bigr),
\qquad p\in\Lambda,
\]
where $x_{\mathrm{ext}}$ equals $x$ on $\Lambda$ and is zero outside $\Lambda$. On the finite image grid, shifting back need not recover samples lost at the boundary.

For an image transformation $g$, let $A_g$ denote the operation that aligns the resulting spatial feature map to the original image coordinates. The main object of interest in our method is
\[
A_g\bigl(\phi(g(x))\bigr).
\]

\noindent\textbf{Perturbations.}
We consider two types of perturbations, extraneous and purposeful. The extraneous one could be an additive perturbation to an image, $\tilde x = x + n$, designed to maximally change the value of the embedding $\phi(\tilde x)$:
\[
n(x)\in\arg\max_{\|\delta\|_p\le\epsilon}
d\bigl(\phi(x),\phi(x+\delta)\bigr),
\qquad
\tilde{x}=x+n(x)
\]
for some small $\epsilon$  designed so the perturbation is, ideally, imperceptible by humans (adversarial perturbation). %

The purposeful perturbations are small translations $g_i, \ i = 1, \dots, N$, which could be sampled deterministically or at random according to some chosen distribution $g_i \sim P_g$, either way yielding a set $
\{g_1(x), \dots, g_N(x)\}$. 
Our goal is to use these purposeful perturbations to combat the effects of extraneous adversarial perturbations.

\noindent\textbf{Smoothing vs. stochastic resonance.}
The resemblance between adversarial perturbations and aliasing has motivated the use of anti-aliasing, or smoothing, techniques to mitigate them. These consist of {\em spatial averaging} of the data prior to computing the map $\phi$. If we call $B^\sigma_{ij}$ a neighborhood of size $\sigma>0$ around $(i,j)\in \Lambda$, 
then the simplest form of smoothing is simply averaging in a neighborhood, 
\[
\bar x_{i,j} = \frac{1}{\sigma^2} \sum_{(i',j') \in B^\sigma_{i,j}}x_{i'j'}
\]
which we write in terms of translations $[g_{u,v}(x)](p)=x_{\mathrm{ext}}(p-(u,v))$ within the same neighborhood $B^\sigma$,
\[
\bar x = \frac{1}{N} \sum_{i=1}^N g_i(x).
\]
One can then obtain an encoding by %
\[
\bar \phi(x) = \frac{1}{N}\sum_{i = 1}^N \phi(g_i(x)).
\]
This can be interpreted as marginalizing over the chosen set of translations with a prior $P_g$ when computing $\phi$. Notice that the average can be computed on a coarser lattice $\tilde \Lambda$, but its value still depends on data in the finer grid $\Lambda$. 
Classical sampling theory teaches that smoothing mitigates the effect of high-frequency aliasing at the cost of information loss on $x$.

Stochastic resonance also marginalizes over the chosen set of translations, but by {\em averaging transformed data in latent space:}
\begin{equation}\vspace{-1mm}
\hat \phi(x) = \frac{1}{N} \sum_{i=1}^N A_{g_i}\bigl(\phi(g_i(x))\bigr).
\label{eq:SR}\vspace{-1mm}
\end{equation}

For an encoder like ViT, $A_{g_i}$ reshapes the spatial tokens of $\phi(g_i(x))$ into a two-dimensional feature map, upsamples it to the input-pixel lattice using nearest-neighbor interpolation, and applies the opposite integer translation on that lattice.
Equation~\eqref{eq:SR} averages these aligned feature maps on the input-pixel lattice, after which we apply average pooling to return to the original token resolution.

\noindent\textbf{Defense against adversarial attacks} (sketch only; please find details in Appendix~\ref{supp:theory}).

Consider a first-order expansion at each transformed input
\[
\phi(g_i\tilde x)=\phi(g_i x + g_i n(x))
\approx \phi(g_i x)+J_{g_i x}\,[g_i n(x)],
\]
where $J_{g_i x}$ is the Jacobian of $\phi$ at $g_i x$. Substituting with \eqref{eq:SR} yields the perturbation term
\begin{equation}\label{eq:sr_expand}
\Delta_{\mathrm{SR}}(x)
=\hat\phi(\tilde x)-\hat\phi(x)
\approx \frac{1}{N}\sum_{i=1}^N A_{g_i}\bigl(J_{g_i x}[g_i n(x)]\bigr).
\end{equation}
By contrast, a standard feedforward encoder responds as
\[
\Delta_{\mathrm{base}}(x)
=\phi(\tilde x)-\phi(x)
\approx J_x\,n(x),
\]
and is therefore dominated by the local worst-case Lipschitz direction at $x$, exploited by the attacker.

The terms $A_{g_i}\bigl(J_{g_i x}[g_i n(x)]\bigr)$ in \eqref{eq:sr_expand} are generally 
\emph{not aligned} across $i$. Under mild assumptions (decorrelation across i, details and empirical evidence in Appendix A),
\[
\mathbb{E}\left\|\frac{1}{N}\sum_{i=1}^N v_i\right\|^2
\approx \frac{1}{N^2}\sum_{i=1}^N \mathbb{E}\|v_i\|^2,
\qquad
v_i = A_{g_i}\bigl(J_{g_i x}[g_i n(x)]\bigr),
\]
provided that the sum of the expected cross terms is small in magnitude
relative to $\sum_{i=1}^N \mathbb{E}\|v_i\|^2$. Under this condition,
the root-mean-square magnitude of the first-order perturbation term
in \eqref{eq:sr_expand} is approximately $1/\sqrt{N}$ times the
root-mean-square magnitude of the individual terms $v_i$ across the
chosen translations $g_i$.

\noindent\textbf{Purposeful perturbation.} The only design choice in this principle is the set of purposeful perturbations. While that can be optimized for performance, we optimize for simplicity, restricting our attention to translation by integer pixels. 
The theory of Stochastic Resonance shows that, if we sub-sample a signal from its native granularity $\Lambda$ to a coarser grid $\tilde \Lambda \subset \Lambda$, and choose the purposeful perturbations to act on the finer grid $\Lambda$, under certain conditions one can recover the original signal at the finer granularity \cite{benzi1981mechanism}. 
Accordingly, we enumerate the bounded integer-pixel translations
\[
\{g_i\}_{i=1}^{N}
=\bigl\{g_{u,v}:(u,v)\in\mathbb{Z}^2,\ |u|\le d_x,\ |v|\le d_y\bigr\},
\qquad N=(2d_x+1)(2d_y+1).
\]
Whenever we report a single SR level $d$, we use $d_x=d_y=d$.

\begin{table*}[t]
\centering
\caption{\small\textbf{Defense against adversarial attacks on CIFAR-10.} Unlike baselines, we perform defense purely at test-time, and achieve state-of-the-art robustness under different attacks. Even in an adaptive setting (worst-case) where the attacker is fully aware of how defense is applied, the effectiveness of adversarial attacks is still notably reduced, while the computational cost for executing such attacks is significantly increased.}
\label{tab:cifar10}

\footnotesize
\setlength{\tabcolsep}{0.8pt}
\begin{tabular}{l|l|>{\centering}p{12mm}|>{\centering}p{12mm}|>{\centering}p{12mm}|>{\centering}p{13mm}|>{\centering}p{12mm}||c||c}
Method & Architecture & Natural & FGSM & PGD-20 & PGD-100 & C\&W & Ensemble & AutoAttack \\
\hline\hline
AT \cite{madry2017towards} & \multirow{8}{*}{ResNet-18} & 85.02 & 56.21 & 48.22 & 46.37 & 47.38 & 45.90 & 44.11 \\
+ FD \cite{xie2019feature} && 85.14 & 56.81 & 48.54 & 46.70 & 47.72 & 45.82 & 44.57 \\
+ CAS \cite{bai2021improving} && \textbf{85.78} & 55.57 & 50.42 & 49.91 & 53.47 & 46.46 & 44.23 \\
+ CIFS \cite{yan2021cifs} && 79.87 & 56.53 & 49.80 & 48.17 & 49.89 & 47.26 & 43.94 \\
+ FSR \cite{kim2023feature} && 81.46 & 58.07 & 52.47 & 51.02 & 49.44 & 48.34 & 46.41 \\
+ TTE \cite{perez2021enhancing} && 85.25 & 59.20 & 53.00 & 51.65 & 52.45 & 50.60 & - \\
\rowcolor{gray!20}
+ Ours && 84.93 & \textbf{61.02} & \textbf{56.08} & \textbf{55.17} & \textbf{55.53} & \textbf{53.68} & \textbf{49.78} \\
\rowcolor{gray!20}
+ Ours (Worst-Case) && 84.93 & 58.81 & 53.58 & 52.39 & 52.73 & 50.95 & - \\
\hline
TRADES \cite{zhang2019theoretically} && 84.92 & 60.87 & 56.13 & 55.16 & 54.02 & 53.38 & 46.81 \\
\rowcolor{gray!20}
+ Ours & \multirow{-2}{*}{WideResNet-34} & \textbf{85.03} & \textbf{62.43} & \textbf{58.64} & \textbf{57.87} & \textbf{57.18} & \textbf{56.28} & \textbf{54.02} \\
\hline
MART \cite{wang2019improving} && \textbf{83.07} & 60.21 & 54.14 & 52.90 & 49.62 & 48.95 & 44.27 \\
\rowcolor{gray!20}
+ Ours & \multirow{-2}{*}{ResNet-18} & 82.70 & \textbf{62.62} & \textbf{59.03} & \textbf{58.13} & \textbf{55.51} & \textbf{54.61} & \textbf{50.72}
\end{tabular}
\end{table*}

\section{Experiments}
\label{sec:experiments}
We evaluate across 6 tasks (image classification, stereo matching, optical flow, semantic segmentation, monocular depth estimation, vision-language alignment), multiple architectures and variants (ResNet, ViT, MLP-Mixer, PSMNet, RAFT), totaling over 100 settings that vary datasets, models, attack types and strengths, and defense strategies. Given the breadth of the evaluation, we focus on the key findings and discussions in the main paper and defer implementation details and additional ablations to the Appendix. %

\subsection{Image Classification}
\label{sec:classification}

\noindent\textbf{CIFAR-10. }
We evaluate on CIFAR-10 \cite{krizhevsky2009learning}, building upon the publicly available code base of FSR \cite{kim2023feature} %
and accompanying evaluation protocol, including AutoAttack \cite{croce2020reliable}. We apply stochastic resonance purely at test-time, to networks pre-trained with AT \cite{madry2017towards}, TRADES \cite{zhang2019theoretically}, and MART \cite{wang2019improving}, using their publicly released weights. We conduct experiments on ResNet-18 \cite{he2016deep} and WideResNet-34 \cite{zagoruyko2016wide}, depending on the availability of author-released pre-trained weights. In all cases, feature ensembling is performed before the final linear layer.
Furthermore, we consider a worst-case adaptive adversary setting, where the attacker has full knowledge of the model weights, and knows every stochastic resonance transformation by accessing \emph{every single} forward pass and its gradients. Implementation details are in Appendix~\ref{supp:classification}.

We compare our method against feature-level manipulation methods (FD \cite{xie2019feature}, CAS \cite{bai2021improving}, CIFS \cite{yan2021cifs}) and ensemble-based approach TTE \cite{perez2021enhancing}. The results (Tab.~\ref{tab:cifar10}), show that stochastic resonance consistently outperforms all baselines across different attacks. Even in the adaptive attack scenario, where the adversary accounts for all stochastic resonance forward passes, the model remains significantly more robust than the baseline methods in non-adaptive settings. Note that, with stochastic resonance, the computational cost for generating adaptive attacks increases substantially: 8$\times$ more wall-clock time is required to create adversarial examples with PGD-100. Fig.~\ref{fig:cifar} shows that increasing the resonance level consistently enhances robustness in all attack settings, leading to substantial improvements over the baseline method (FSR).%

\begin{wraptable}{r}{0.5\textwidth}
\vspace{-3mm}
\setlength{\tabcolsep}{0.8pt} 
  \captionof{table}{\small{\bf ImageNet with ViT-Small.}  Increasing the level of stochastic resonance consistently improves both Top-1 and Top-5 accuracy under adversarial attacks. Relative to the clean baseline (72.9 Top-1, 92.91 Top-5), our method recovers up to 55.8\% of the Top-1 accuracy loss and 68.1\% of the Top-5 accuracy loss.}
  \label{tab:ViT}
\scriptsize
  \centering
\resizebox{1\linewidth}{!}{
  \begin{tabular}{l|c|c|c|c|c|c|c|c}
Att. Strength ($\epsilon$)&\multicolumn{2}{c|}{8/255}&\multicolumn{2}{c|}{4/255}&\multicolumn{2}{c|}{2/255}&\multicolumn{2}{c}{1/255}\\
\hline
Metric&Top-1&Top-5&Top-1&Top-5&Top-1&Top-5&Top-1&Top-5\\\hline
No Defense&4.51&19.55&12.25&42.15&29.79&65.09&48.09&78.13\\
Ours ($d=1$)&11.66&46.27&27.85&65.80&45.94&77.63&57.48&83.43\\
Ours ($d=2$)&18.78&58.58&36.42&72.86&51.58&80.85&60.41&84.77\\
Ours ($d=3$)&\textbf{25.77}&\textbf{65.88}&\textbf{42.52}&\textbf{76.75}&\textbf{54.94}&\textbf{82.64}&\textbf{62.08}&\textbf{85.50}
    \end{tabular} 
    } 
     \vspace{-2mm}
\end{wraptable}
\noindent\textbf{ImageNet.}
We further evaluate our approach on the ImageNet \cite{deng2009imagenet} using a ViT \cite{dosovitskiy2020image} backbone. We vary the level of stochastic resonance and conduct ablation studies by testing against PGD attacks of different strengths. Tab.~\ref{tab:ViT} reports the results, where increasing the resonance level leads to consistent improvements in adversarial robustness. Notably, our method recovers the accuracy drop under adversarial attacks by up to a relative 55.8\% (Top-1, when $\epsilon=4/255$) and 68.1\% (Top-5, when $\epsilon=2/255$). Our experiments on ResNet-50 also demonstrate a consistent improvement, as in Tab.~\ref{tab:layers} and Tab.~\ref{tab:Rotation_supp} in the Appendix.

\def\figd{Figures/Cifar}
\def\fWidD{0.2\textwidth}
\begin{figure*}[t]

\centering
\hspace*{-4mm}
{\scriptsize
\begin{tabular}[0mm]{c@{\hskip 0.01in}c@{\hskip 0.01in}c@{\hskip 0.01in}c@{\hskip 0.01in}c}
\includegraphics[width=\fWidD]{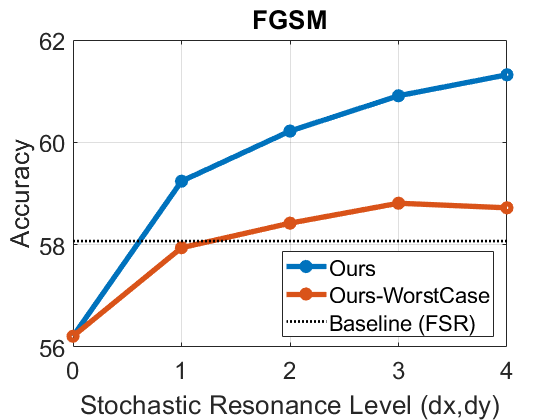}&
\includegraphics[width=\fWidD]{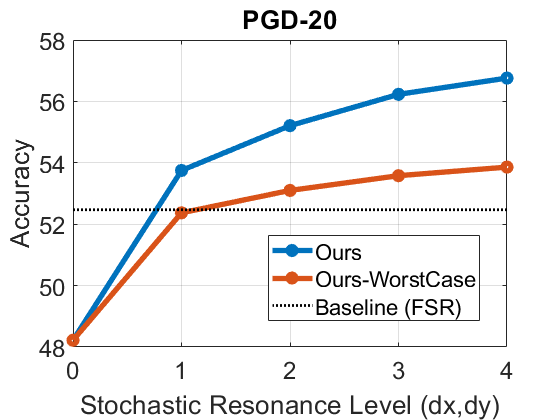}&
\includegraphics[width=\fWidD]{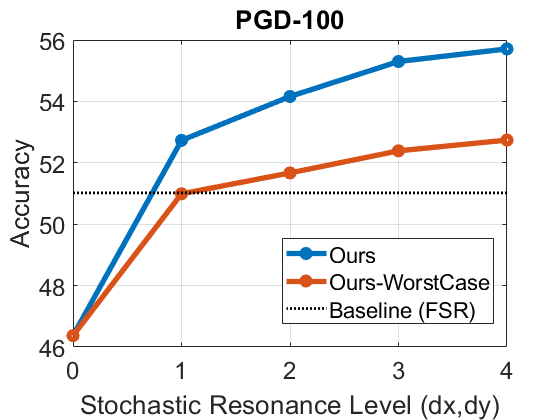}&
\includegraphics[width=\fWidD]{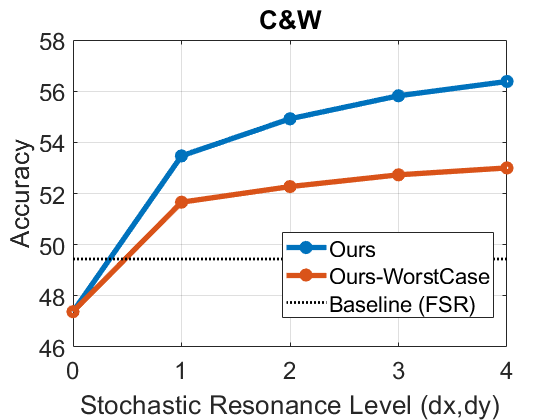}&
\includegraphics[width=\fWidD]{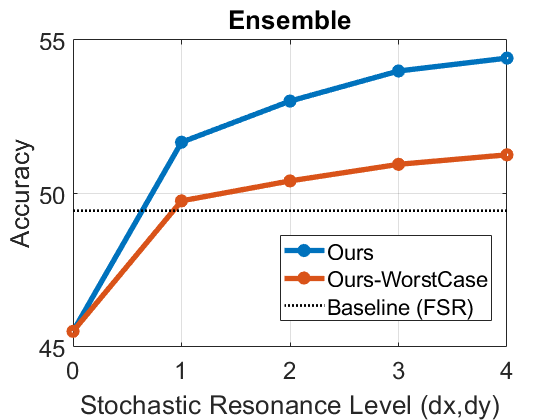}
\end{tabular}
}\vspace{-2mm}
\caption{\small {\bf Varying levels of stochastic resonance.} Increasing the stochastic resonance level enhances robustness across all settings. Notably, our approach achieves superior performance even under adaptive attacks (Ours-WorstCase), despite the baseline (FSR) being evaluated only in the non-adaptive case.}

\label{fig:cifar}
\end{figure*}

\noindent\textbf{Applying stochastic resonance to early layers.} (Details in Appendix~\ref{supp:layer}.)
Since adversarial perturbations often appear as high-frequency noise, early feature extraction is a natural point of intervention. As shown in Tab.~\ref{tab:layers}, applying stochastic resonance through only the first residual block of ResNet-50 already provides strong robustness, while extending it to the second block yields the best performance. This suggests that running stochastic resonance through shallow layers can be sufficient for defense, substantially reducing computational overhead.

\noindent\textbf{Alternative purposeful perturbations.} (Details in Appendix~\ref{supp:rotation}.)
We also evaluate rotations as an alternative family of purposeful perturbations. As shown in Tab.~\ref{tab:rotation_alignment}, rotations improve adversarial robustness across all resonance levels, with larger gains at lower resonance levels. We hypothesize that this is partly due to the implementation of standard libraries. Unlike translations with integer displacements, rotations typically map pixel coordinates to non-integer locations and therefore require interpolation-based resampling, which can introduce artifacts such as aliasing and blurring. Higher-fidelity resampling strategies, such as supersampling or Fourier-domain rotation, may mitigate these artifacts. Nevertheless, the robustness gains achieved by rotations provide a proof of concept: our method extends beyond translations and is not specific to a single transformation class.

\noindent\textbf{Extending to MLP-based architectures.} (Details in Appendix~\ref{supp:mlpmixer}.) We further check whether stochastic resonance extends beyond CNNs and Transformers by evaluating it on MLP-Mixer~\cite{tolstikhin2021mlp}. On CIFAR-10, stochastic resonance preserves clean accuracy while substantially improving adversarial robustness: under PGD with $\epsilon=8/255$, accuracy increases from $0.03$ without defense to $65.38$ with stochastic resonance at $d=3$. Similar gains are observed under FGSM, where accuracy improves from $21.76$ to $74.08$ at $\epsilon=8/255$.

\noindent\textbf{Real-world corruptions}. We also test how stochastic resonance behaves under real-world corruptions. To this end, we evaluate on CIFAR-10-C~\cite{hendrycksbenchmarking}, which contains a diverse set of corruptions covering noise, blur, weather effects, and photometric changes. With ResNet-50, following the best configuration identified in prior sections, stochastic resonance recovers an average of 20.8\% of lost accuracy, including 32.6\% under \texttt{defocus\_blur}, 34.4\% under \texttt{fog}, 31.6\% under \texttt{gaussian\_blur}, and 31.9\% under \texttt{saturate}. Other corruptions also see recovery ranging from 3.4\% to 20.8\%. These findings highlight a key strength of stochastic resonance: it works across a wide spectrum of \emph{unknown} corruptions, strengthening it as a \emph{general-purpose}, \emph{plug-and-play} defense.%

\begin{figure*}[t]
\centering
\includegraphics[width=\textwidth]{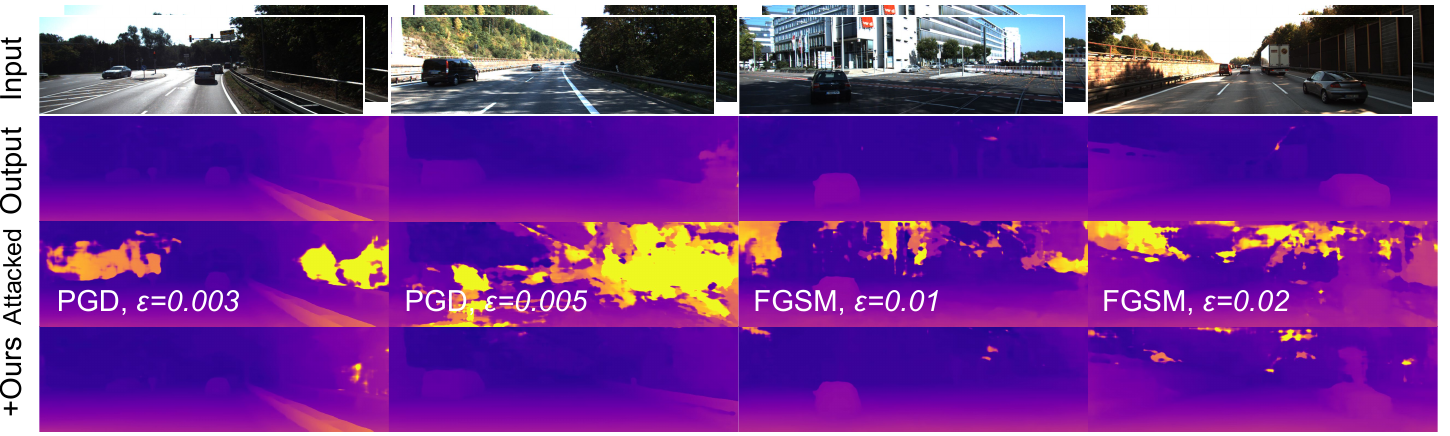}
    \caption{\small {\bf Qualitative results on stereo matching.} %
    We demonstrate a significant reduction in prediction errors. This technique holds significant potential for safety-critical applications where stereo vision must remain reliable under diverse environmental conditions and adversarial threats. Additional visualizations at Appendix~\ref{supp:vis}.
}
    \label{fig:stereo}
\end{figure*}

\begin{table*}[t]
\setlength{\tabcolsep}{0.8pt} 
  \caption{\small{\bf Stochastic resonance enhances stereo matching robustness.} Stochastic resonance significantly mitigates errors across all settings, reducing error by up to 71.89\% (MAE, attacked by PGD with $\epsilon=0.005$).} 
\footnotesize
  \centering
  \begin{tabular}{c|l|c|c|c|c|c|c|c|c|c|c|c|c}
  &Attack Strength ($\epsilon$)&\multicolumn{3}{c|}{0.02}&\multicolumn{3}{c|}{0.01}&\multicolumn{3}{c|}{0.005}&\multicolumn{3}{c}{0.002}\\\cline{2-14}
&Metric&\,MAE\,&RMSE&D1-err&\,MAE\,&RMSE&D1-err&\,MAE\,&RMSE&D1-err&\,MAE\,&RMSE&D1-err\\\hline
{\multirow{5}{*}{\rotatebox[origin=c]{90}{FGSM}}} &No Defense&14.83&24.10&97.33&8.49&14.53&90.49&5.05&7.70&74.71&3.01&3.49&38.12 \\
&Latent Smoothing~\cite{addepalli2021boosting}&13.42&22.09&96.61&8.12&13.69&89.25&4.89&7.05&73.01&2.89&3.32&36.25
\\
&Randomized Smoothing~\cite{cohen2019certified}&14.20&22.41&97.26&7.38&11.21&89.44&4.26&5.19&70.49&2.78&3.18&30.76 \\
&Ours ($d=1$)&10.12&15.81&95.80&6.46&9.92&86.30&4.39&5.93&68.55&2.78&3.18&31.99 \\
&Ours ($d=2$)&\textbf{9.22}&\textbf{13.88}&\textbf{94.61}&\textbf{6.13}&\textbf{8.87}&\textbf{84.49}&\textbf{4.19}&\textbf{5.43}&\textbf{66.74}&\textbf{2.73}&\textbf{3.12}&\textbf{31.18}\\
&\cellcolor{gray!20}Error Reduced (\%)&\cellcolor{gray!20} 37.82&\cellcolor{gray!20}42.40&\cellcolor{gray!20}2.79&\cellcolor{gray!20}27.78&\cellcolor{gray!20}38.95&\cellcolor{gray!20}6.63&\cellcolor{gray!20}16.92&\cellcolor{gray!20}29.41&\cellcolor{gray!20}10.66&\cellcolor{gray!20}9.21&\cellcolor{gray!20}10.65&\cellcolor{gray!20}18.21 \\\hline
{\multirow{5}{*}{\rotatebox[origin=c]{90}{PGD}}} &No Defense&161.70&162.61&99.99&131.66&140.55&98.64&63.97&88.55&85.31&6.83&17.03&39.24 \\
&Latent Smoothing~\cite{addepalli2021boosting}&161.79&162.69&99.99&131.46&140.41&98.57&63.86&88.21&85.39&7.28&17.44&39.81\\
&Randomized Smoothing~\cite{cohen2019certified}&149.00&153.36&99.95&62.59&83.94&95.09&11.06&19.03&69.85&2.69&3.38&25.01\\
&Ours ($d=1$)&107.86&125.14&98.29&69.97&84.72&91.79&20.66&42.29&73.21&4.17&8.32&29.09\\

&Ours ($d=2$)&\textbf{77.59}&\textbf{100.14}&\textbf{96.14}&\textbf{44.77}&\textbf{66.23}&\textbf{89.24}&\textbf{17.98}&\textbf{32.80}&\textbf{71.09}&\textbf{3.76}&\textbf{6.73}&\textbf{28.44}\\
&\cellcolor{gray!20}Error Reduced (\%)&\cellcolor{gray!20} 52.01 &\cellcolor{gray!20} 38.41 &\cellcolor{gray!20} 3.85 &\cellcolor{gray!20} 66.02 &\cellcolor{gray!20} 52.87 &\cellcolor{gray!20} 9.53 &\cellcolor{gray!20} 71.89 &\cellcolor{gray!20} 62.94 &\cellcolor{gray!20} 16.66 &\cellcolor{gray!20} 44.99 &\cellcolor{gray!20} 60.52 &\cellcolor{gray!20} 27.53
\end{tabular}
     \label{tab:stereo}
\end{table*}

\subsection{Dense Prediction}
\label{sec:dense}
We next evaluate stochastic resonance on dense prediction tasks. This setting highlights a key strength of our method: standard data augmentations are often infeasible because they can alter the input geometry and violate the underlying physics of the task. In contrast, stochastic resonance ensembles features in latent space after alignment, keeping predictions registered to the original input.

\noindent\textbf{Stereo matching. } (Details in Appendix~\ref{supp:stereo}.) 
Stereopagnosia \cite{berger2022stereoscopic,wong2021stereopagnosia} first introduced adversarial attacks to stereo matching, yet no test-time method has demonstrated an effective defense. We evaluate our method on the standard benchmark used in \emph{Stereopagnosia}, derived from KITTI~\cite{geiger2012we}. Experiments are conducted using FGSM and PGD attacks against a pre-trained PSMNet~\cite{chang2018pyramid}. Because no established test-time defense is available, we implement latent-space smoothing~\cite{addepalli2021boosting} and randomized smoothing~\cite{cohen2019certified} as baseline methods. As shown in Tab.~\ref{tab:stereo}, both attacks corrupt network predictions, with PGD proving more effective due to its iterative nature. Nevertheless, stochastic resonance consistently improves robustness across attack strengths. In particular, under PGD with $\epsilon=0.005$, our method reduces errors by up to 71.89\% in terms of MAE.
Fig.~\ref{fig:stereo} provides qualitative results. The attacks introduce significant errors, as indicated by bright regions in the predictions. Stochastic resonance effectively mitigates these errors. Note that this is achieved with neither prior knowledge of the attack nor additional robust training. This experiment is particularly relevant for safety-critical applications such as autonomous driving, where adversarial disturbances can arise not only from malicious attacks but also from environmental factors such as adverse weather conditions, varying illumination, or sensor degradation. While some  defenses \cite{zhang2023weatherstream, cheng2021towards, cheng2022revisiting, berger2022stereoscopic} have been proposed to train a more robust network under adverse conditions, test-time defenses remain largely unexplored. %

\def\figd{Figures/RAFT}
\def\fWidD{0.24\textwidth}
\begin{wrapfigure}{r}{0.5\textwidth}
\vspace{-1mm}
\centering
{
\footnotesize
\includegraphics[width=\fWidD]{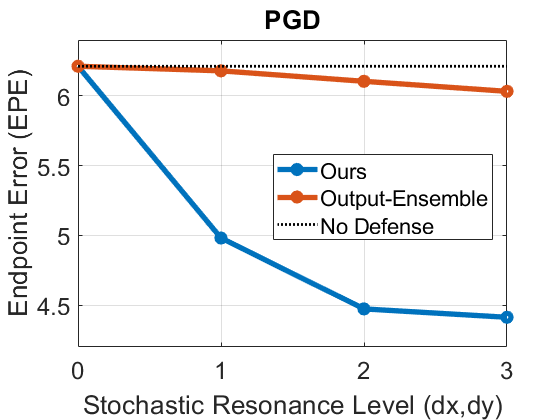}
\includegraphics[width=\fWidD]{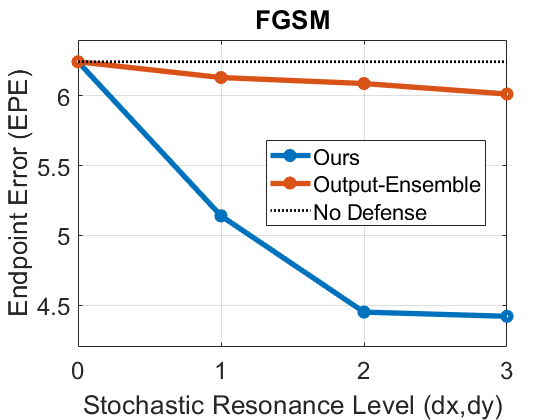} 
\caption{\small {\bf Improved optical flow robustness.} By ensembling in the latent feature space rather than the output space, stochastic resonance consistently achieves stronger robustness across all resonance levels.}
\label{fig:flow-quantitative}
}

\end{wrapfigure}

\noindent\textbf{Optical flow. } (Details in Appendix~\ref{supp:optical-flow}.) We further evaluate our method on optical flow, which computes the dense motion field between two images. The accuracy is measured using the End-Point Error (EPE). Although several adversarial attacks have been proposed \cite{schrodi2022towards, Ranjan_2019_ICCV, schmalfuss2022perturbation, agnihotri2023cospgd, scheurer2024detection}, there is currently no standardized benchmark for the evaluation of defense. Therefore, we adopt a simple and controlled setup: we employ the RAFT \cite{teed2020raft} %
model and focus on global perturbations generated by PGD and FGSM. We apply stochastic resonance to the feature extractor of RAFT, introducing no overhead to the computationally intensive matching module. We test on the DAVIS \cite{pont20172017} dataset. As in Fig.~\ref{fig:flow-quality}, our approach effectively removes errors caused by adversarial noise. Quantitative results (Fig.~\ref{fig:flow-quantitative}) show that increasing stochastic resonance reduces EPE, which aligns with our findings in classification. We also compare our method to an alternative ensembling approach that aggregates predictions in the output space, similar to TTE \cite{perez2021enhancing}. We apply the same stochastic transformations, but instead perform an ensemble at the output level (with alignment). This approach remains less effective than our approach. This finding highlights the advantage of having the freedom to ensemble at any layer.

\def\figd{Figures/Flow}
\def\fWidD{0.14\textwidth}
\begin{figure*}[t]

\renewcommand{\arraystretch}{0.2} %
\centering
\hspace*{-4mm}
{\scriptsize
\begin{tabular}[0mm]{c@{\hskip 0.01in}c@{\hskip 0.01in}c@{\hskip 0.01in}c|c@{\hskip 0.01in}c@{\hskip 0.01in}c}
& \multicolumn{3}{c}{\normalsize PGD}& \multicolumn{3}{c}{\normalsize FGSM}\\\\
\smash{\rotatebox{90}{\quad Image}}&
\includegraphics[width=\fWidD]{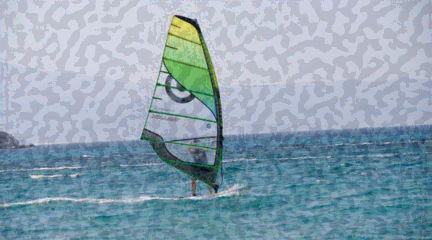}&
\includegraphics[width=\fWidD]{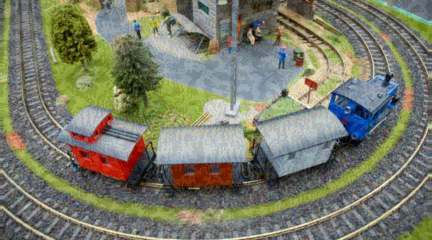}&
\includegraphics[width=\fWidD]{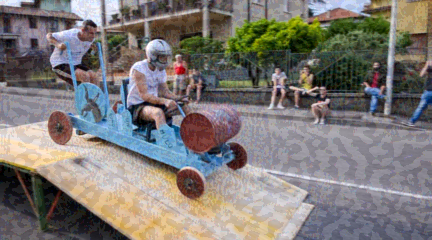}&
\includegraphics[width=\fWidD]{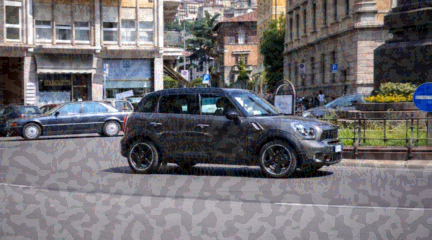}&
\includegraphics[width=\fWidD]{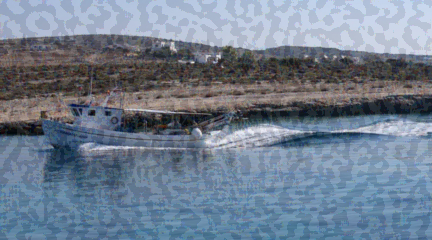}&
\includegraphics[width=\fWidD]{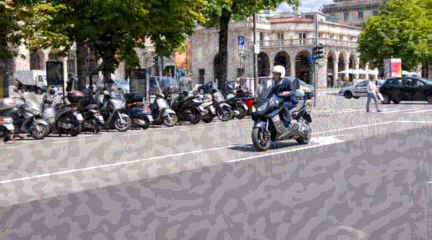}\\
\smash{\rotatebox{90}{Original}}&
\includegraphics[width=\fWidD]{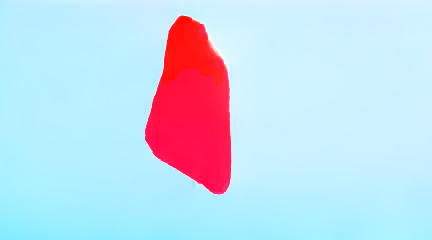}&
\includegraphics[width=\fWidD]{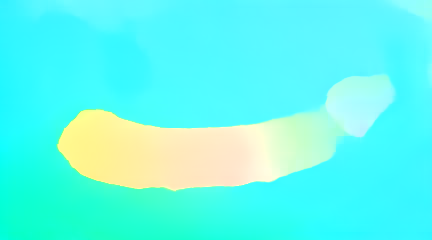}&
\includegraphics[width=\fWidD]{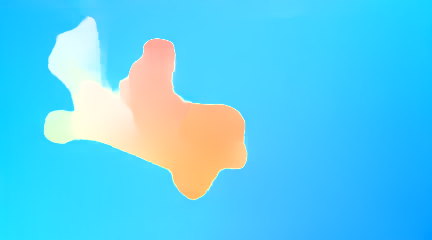}&
\includegraphics[width=\fWidD]{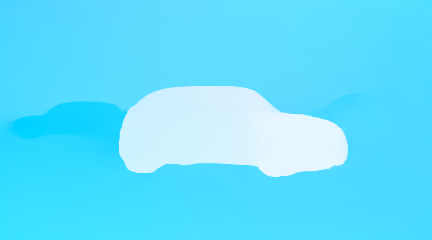}&
\includegraphics[width=\fWidD]{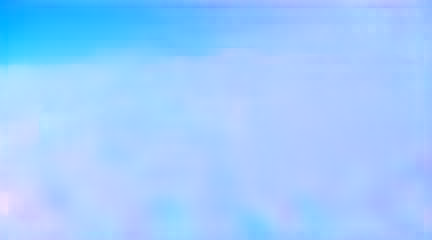}&
\includegraphics[width=\fWidD]{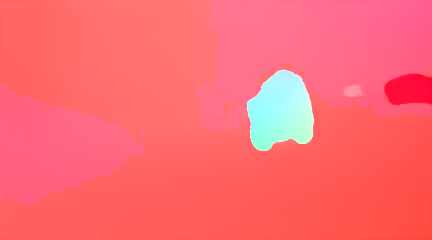}\\
\smash{\rotatebox{90}{Attacked}}&
\includegraphics[width=\fWidD]{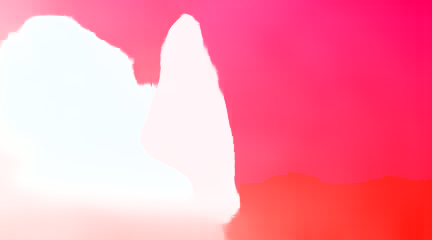}&
\includegraphics[width=\fWidD]{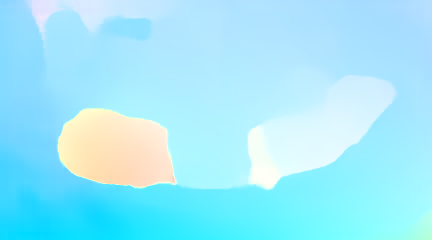}&
\includegraphics[width=\fWidD]{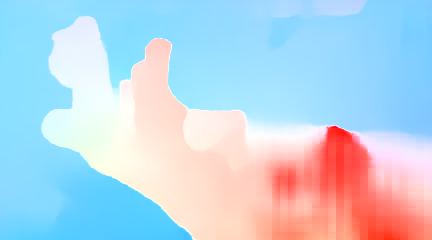}&
\includegraphics[width=\fWidD]{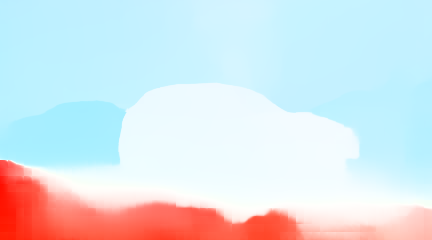}&
\includegraphics[width=\fWidD]{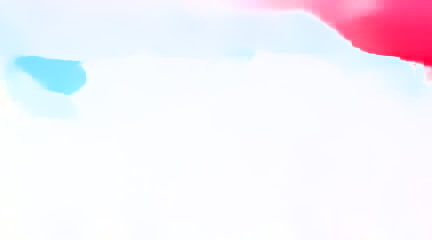}&
\includegraphics[width=\fWidD]{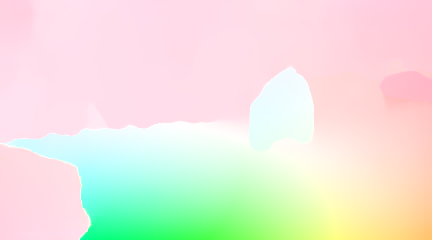}\\
\smash{\rotatebox{90}{\quad Ours}}&
\includegraphics[width=\fWidD]{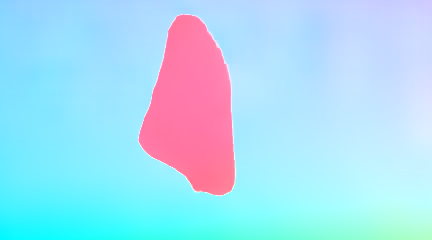}&
\includegraphics[width=\fWidD]{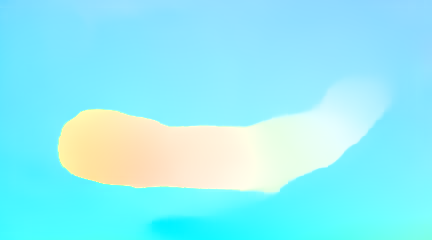}&
\includegraphics[width=\fWidD]{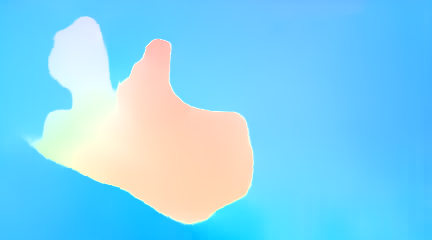}&
\includegraphics[width=\fWidD]{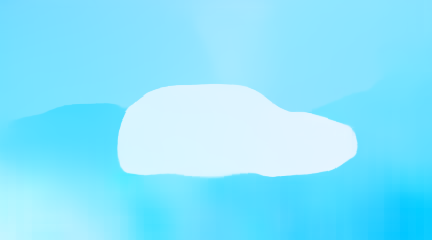}&
\includegraphics[width=\fWidD]{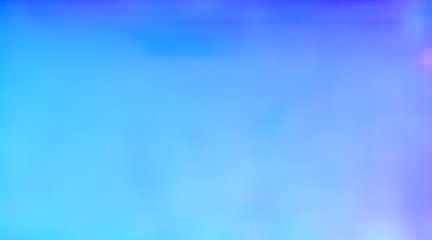}&
\includegraphics[width=\fWidD]{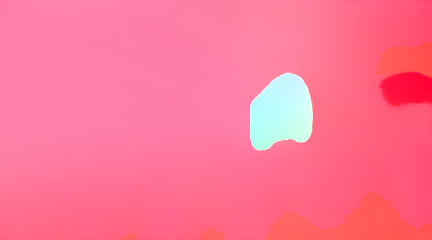}\\
\end{tabular}
}
\caption{\small{\bf Qualitative results for optical flow.} 
Our method substantially mitigates the degradation caused by both PGD and FGSM attacks. Such robustness is critical for perception systems that rely on motion estimation.}
\label{fig:flow-quality}
\end{figure*}

We further evaluate stochastic resonance under adverse weather conditions~\cite{schmalfuss2023distracting}. This setting poses two unique challenges: (i) the corruptions are \emph{not} optimized for the model and instead reflect natural image statistics, and (ii) the perturbations violate the underlying theoretical assumptions in Sec.~\ref{sec:method}. Despite these challenges, stochastic resonance still recovers 7.5\% of the accuracy loss for \texttt{snow} and 3.7\% for \texttt{rain}. While, as expected, the gains are smaller than observed under adversarial attacks, they nonetheless demonstrate that stochastic resonance provides meaningful robustness against real-world perturbations, consistent with our findings on CIFAR-10-C. %

\noindent\textbf{Semantic segmentation.}
We next evaluate stochastic resonance on semantic segmentation, where the model assigns a class label to each pixel and must localize the boundaries between adjacent regions. We use Cityscapes~\cite{cordts2016cityscapes} and evaluate under CosPGD~\cite{agnihotri2023cospgd} and SegPGD~\cite{gu2022segpgd} at three perturbation strengths, $\epsilon\in\{2,4,8\}/255$. We follow their protocols, except for adding Stochastic Resonance at test time. In addition to mean intersection over union (mIoU), we report boundary mIoU to examine whether the gains from aggregating aligned features extend to pixels near object boundaries. As shown in Tab.~\ref{tab:cityscapes_sr}, both metrics improve with each increase in the resonance level across all tested attack strengths. For example, under CosPGD with $\epsilon=4/255$, stochastic resonance at $d=3$ increases mIoU from 17.16 to 27.38 and boundary mIoU from 13.74 to 23.28. SegPGD yields lower scores at the same perturbation strength, but we observe a similar trend: mIoU increases from 0.47 to 9.06 and boundary mIoU from 0.49 to 9.32. At the largest SegPGD budget, $\epsilon=8/255$, the gains are smaller, with mIoU reaching 2.30 at $d=3$.

\begin{table}[t]
\centering
\footnotesize
\setlength{\tabcolsep}{4pt}
\caption{\textbf{Semantic segmentation on Cityscapes under attack.}
We report mIoU and boundary mIoU (\%) for the baseline model and three
levels of stochastic resonance. Higher is better for both metrics.}
\label{tab:cityscapes_sr}
\begin{tabular}{@{}lc*{8}{r}@{}}
\toprule
& & \multicolumn{4}{c}{mIoU $\uparrow$}
  & \multicolumn{4}{c}{Boundary mIoU $\uparrow$} \\
\cmidrule(lr){3-6}\cmidrule(l){7-10}
Attack & $\epsilon$
& No SR & $d=1$ & $d=2$ & $d=3$
& No SR & $d=1$ & $d=2$ & $d=3$ \\
\midrule
CosPGD & $2/255$ & 18.40 & 25.53 & 31.14 & \textbf{32.75}
                 & 15.14 & 21.57 & 26.71 & \textbf{28.21} \\
       & $4/255$ & 17.16 & 22.42 & 26.29 & \textbf{27.38}
                 & 13.74 & 18.61 & 22.26 & \textbf{23.28} \\
       & $8/255$ & 15.39 & 19.15 & 21.71 & \textbf{22.41}
                 & 12.18 & 15.57 & 17.96 & \textbf{18.62} \\
\midrule
SegPGD & $2/255$ &  1.02 &  5.64 & 18.24 & \textbf{23.46}
                 &  1.08 &  6.06 & 17.84 & \textbf{22.11} \\
       & $4/255$ &  0.47 &  1.50 &  6.43 & \textbf{9.06}
                 &  0.49 &  1.66 &  6.79 & \textbf{9.32} \\
       & $8/255$ &  0.40 &  0.67 &  1.70 & \textbf{2.30}
                 &  0.41 &  0.72 &  1.81 & \textbf{2.41} \\
\bottomrule
\end{tabular}
\end{table}

\begin{wraptable}{r}{0.61\textwidth}
\vspace{-2mm}
\setlength{\tabcolsep}{2.0pt}
\caption{\small{\bf Monocular depth on NYU-V2 with larger ViT backbones.} We evaluate under PGD-10 attacks with $\epsilon=8/255$. Error reduction is measured as relative recovery toward the clean baseline.}
\footnotesize
\centering
\begin{tabular}{l|l|c|c|c|c}
Backbone & Method & RMSE $\downarrow$ & RMSE-log $\downarrow$ & AbsRel $\downarrow$ & SqRel $\downarrow$ \\
\hline
\multirow{4}{*}{ViT-B/14} 
& Clean & 0.396 & 0.135 & 0.100 & 0.061 \\
& No Defense & 0.645 & 0.191 & 0.168 & 0.383 \\
& Ours & 0.564 & 0.181 & 0.155 & 0.163 \\
& \cellcolor{gray!20}Error Reduced (\%) 
& \cellcolor{gray!20}32.5
& \cellcolor{gray!20}17.9
& \cellcolor{gray!20}19.1
& \cellcolor{gray!20}68.3 \\
\hline
\multirow{4}{*}{ViT-L/14}
& Clean & 0.373 & 0.127 & 0.093 & 0.054 \\
& No Defense & 0.588 & 0.189 & 0.163 & 0.236 \\
& Ours & 0.573 & 0.186 & 0.159 & 0.150 \\
& \cellcolor{gray!20}Error Reduced (\%) 
& \cellcolor{gray!20}7.0
& \cellcolor{gray!20}4.8
& \cellcolor{gray!20}5.7
& \cellcolor{gray!20}47.3
\end{tabular}

\label{tab:nyuv2_depth}
\vspace{-3mm}
\end{wraptable}
\noindent\textbf{Monocular depth estimation.} (Details in Appendix~\ref{supp:mono-depth-est}.)
We also evaluate stochastic resonance on monocular depth estimation with larger ViT backbones. We use DINOv2~\citep{oquab2023dinov2} ViT-B/14 and ViT-L/14 backbones on NYU-V2 \citep{Silberman:ECCV12}. Under PGD-10 attacks with $\epsilon=8/255$, stochastic resonance consistently recovers performance across all evaluation metrics. The strongest recovery is observed on SqRel, where stochastic resonance recovers $68.3\%$ of the attack-induced degradation for ViT-B/14 and $47.3\%$ for ViT-L/14. These results further indicate that the method remains effective for larger transformer backbones and higher-dimensional embeddings.

\subsection{Vision-Language Alignment}
\label{sec:vlm}

\begin{wraptable}{r}{0.5\textwidth}
\setlength{\tabcolsep}{3.2pt}
\caption{\small{\bf Vision-language alignment under attack.} On CLIP ViT-L/14 under PGD attacks, stochastic resonance improves image-text alignment as the resonance level increases, and complements adversarial training (AT).}

\scriptsize
\centering
\resizebox{1\linewidth}{!}{
\begin{tabular}{l|c|c|c|c|c}
Defense & Clean & Baseline & $d=1$ & $d=2$ & $d=3$ \\
\hline
No Defense & 0.302 & -0.025 & 0.064 & 0.145 & \textbf{0.179} \\
AT & 0.302 & 0.136 & 0.195 & 0.231 & \textbf{0.249}
\end{tabular}
}\vspace{-1mm}

\label{tab:clip}

\end{wraptable}
We further evaluate whether stochastic resonance remains effective for vision-language models. We use CLIP~\citep{clip} ViT-L/14 and evaluate image-text alignment on COCO-Captions \citep{lin2014microsoft} under PGD attack, using cosine similarity between image and text embeddings as the objective. (Details in Appendix~\ref{supp:vlm}.) The attack substantially the image-text alignment, reducing cosine similarity from $0.302$ to $-0.025$. Stochastic resonance progressively restores the alignment as the resonance level increases, reaching $0.179$ with $d=3$ (Tab.~\ref{tab:clip}). We also evaluate adversarial training (AT). Although AT is not directly comparable to our training-free method, stochastic resonance outperforms it under the same evaluation and remains complementary: combining adversarial training with stochastic resonance yields the strongest result.

We also evaluate whether stochastic resonance complements existing CLIP defenses on zero-shot ImageNet classification. Following the TTC protocol~\cite{xing2025clip}, we evaluate CLIP ViT-B/32 under PGD-10 with $\epsilon=1/255$. At $d=3$, adding SR to TTC increases adversarial Top-1/Top-5 accuracy (\%) from 38.49/63.12 to 48.10/75.09, while AOM~\cite{tong2025zero} achieves 44.06/68.98 under the same protocol. For CSR~\cite{nie2026contrastive}, applying SR to classify the rectified image increases adversarial Top-1/Top-5 accuracy from 18.76/38.62 to 31.08/63.54. Adding SR also improves clean Top-1 accuracy for both TTC (49.34\% to 54.92\%) and CSR (58.65\% to 59.43\%).

Note that, in modern vision-language models, the vision encoder is typically a small component relative to the language model. For example, in LLaVA-13B \citep{liu2023visual}, the vision encoder occupies roughly $1.1$GB in fp32, whereas the language backbone requires more than $24$GB even in fp16. Since stochastic resonance is applied only to the vision encoder, the added cost is limited to this smaller component, increasing less than $10\%$ additional memory for models like LLaVA-13B in our best settings.%
This makes our method practical for current VLMs, where robustness can be improved without modifying or repeatedly evaluating the substantially larger language backbone.

\section{Discussion and Conclusion}\label{sec:discussion}

\noindent\textbf{Speed.} 
Stochastic resonance incurs low overhead due to parallel computing: raising $d$ to 3 with ResNet-50 adds only 0.06 seconds to the inference time on an NVIDIA 1080Ti GPU. Notably, most of the overhead arises from transformations implemented by Python loops, which is expected to be drastically reduced with efficient CUDA implementations. Excluding this \texttt{for} loop, forward computation is lightweight: 0.028s with $d{=}3$ (49 forward passes), while a single forward pass ($d{=}1$) takes 0.007s. In Appendix~\ref{supp:trade-off} we provide a detailed speed-performance tradeoff.
Moreover, robustness can be attained by applying stochastic resonance only through shallow layers, offering greater efficiency than existing ensemble-based defenses (e.g. \cite{perez2021enhancing}). Moreover, stochastic resonance is fully plug-and-play. In contrast, adversarial training is attack-specific, over 6× slower than standard training, and must be repeated for each type of threat. As such, the computation of our method is well justified by its robustness gains and training-free nature.

\noindent\textbf{Memory.} Increasing $d$ from 0 to 3 raises the number of forward passes from 1 to 49, while peak memory grows only moderately (158MB to 650MB) for ImageNet-sized inputs on ResNet-50 (details in Appendix~\ref{supp:memory}), which is easily handled by modern GPUs. In contrast, memory usage for \emph{adaptive} attack jumps from 346MB to 10.6GB for a single 224$\times$224 image, rendering such attacks impractical on standard hardware and conferring an inherent defensive advantage. Overall, stochastic resonance provides a favorable cost-robustness trade-off with modest memory growth.

\noindent\textbf{Limitations and future work. } Although the computation can be parallelized, the overhead may still be non-negligible in settings with strict memory, latency, or power constraints. More broadly, how to dynamically allocate test-time compute remains an open problem: different inputs, attacks, and deployment environments may require different resonance levels and applying stochastic resonance to different layers. Selecting this budget adaptively is an important direction for future work.

\noindent\textbf{Conclusion. }
We present a signal-processing perspective on adversarial robustness that connects adversarial perturbations to aliasing effects, and propose stochastic resonance as a principled ``combat noise with noise'' test-time defense. Implemented through simple pixel-level translations and inverse feature alignment, our approach is training-free, architecture-agnostic, and broadly applicable across tasks and attack models. Extensive experiments demonstrate state-of-the-art robustness on image classification benchmarks, including under adaptive adversaries, and establish the first effective test-time defense for dense prediction tasks, achieving up to a 71\% error reduction in stereo matching. Beyond strong empirical performance, our results highlight stochastic resonance as a simple yet powerful mechanism for improving robustness without additional training, offering a practical path toward more reliable deployment of models in adversarial and real-world settings.

\begin{ack}
Dong Lao is supported by start-up grants from Louisiana State University.
Alex Wong is supported by NSF under the NSF AI Institute for Edge Computing and Systems (Athena), Grant No. 2112562.
The authors have no additional competing interests to disclose.
\end{ack}

\clearpage

\bibliography{main}
\bibliographystyle{plain}

\clearpage
\appendix

\section{Additional Theoretical Foundation for Stochastic Resonance}\label{supp:theory}

The theoretical foundations of Stochastic Resonance have been long established in statistical physics, summarised for instance in \cite{gammaitoni1998stochastic}
. Our work applies the concept to trained embeddings, but we note that, in the context of adversarial perturbations, modern deep networks often exhibit poor local Lipschitz regularity, where small input perturbations induce disproportionately large latent-space changes, exploited by adversaries.

If $x$ is a clean signal (an image in our case) and $\tilde{x} = x + n(x)$ its adversarially perturbed counterpart. Let $\varphi$ denote the encoder up to the SR layer. Using a discretized set of translations $ \lbrace g_i \rbrace^N_{i=1} \subset G $ and their corresponding inverse $ g^{-1}_i$, the SR feature is:

$$
\hat{\varphi}(x)=\frac{1}{N}\sum_{i=1}^N g^{-1}_{i}\varphi(g_i x), \qquad \hat{\varphi}(\tilde{x})=\frac{1}{N} \sum_{i=1}^N g^{-1}_{i}\varphi(g_i \tilde{x})
$$

The first-order approximation yields:

$$
\varphi(g_i \tilde{x}) = \varphi(g_i x + g_i n(x))
\approx \varphi(g_i x) + J_{g_i x}[g_i n(x)],
$$

where $J_{g_i x}$ is the Jacobian of $\varphi$ at $g_i x$. We have

$$
\Delta_{\mathrm{SR}}(x)
=\hat{\varphi}(\tilde{x})-\hat{\varphi}(x)
\approx \frac{1}{N}\sum_{i=1}^N g^{-1}_{i} J_{g_i x} [g_i n(x)].
$$

In contrast, a standard encoder responds as:

$$
\Delta_{\mathrm{base}}(x)
=\varphi(\tilde{x})-\varphi(x)
\approx J_x n(x),
$$

Here $J_x n(x)$ is the first-order response of the encoder to the
perturbation $n(x)$. For stochastic resonance, the corresponding terms
$v_i=A_{g_i}\bigl(J_{g_i x}[g_i n(x)]\bigr)$ in
\eqref{eq:sr_expand} arise from the same image perturbation, evaluated
under different translations and aligned to the original image coordinates.
Taking the expectation over inputs and their corresponding adversarial
perturbations gives
\[
\mathbb{E}\left\|\frac{1}{N}\sum_{i=1}^N v_i\right\|^2
=
\frac{1}{N^2}\left(
\sum_{i=1}^N \mathbb{E}\|v_i\|^2
+
\sum_{i\ne j}\mathbb{E}\langle v_i,v_j\rangle
\right).
\]
If the second sum is small in magnitude relative to the first, the
root-mean-square magnitude of the averaged first-order response is
approximately $1/\sqrt{N}$ times the root-mean-square magnitude of
the individual terms $v_i$ across the chosen translations. This
interpretation assumes that the local first-order expansion is accurate
at the perturbation magnitude under consideration. A worst-case
Lipschitz bound would require control over all perturbation directions.

To examine alignment empirically, we report the mean pairwise correlation
of the aligned perturbation responses at each layer of a pre-trained
ViT-S on ImageNet under adaptive PGD-100, where the attack differentiates
through all translations (Table~\ref{tab:vit_corr}).

\begin{table}[h]
\caption{\small Mean pairwise correlation of the aligned perturbation terms $v_i$ across layers of a ViT-S model under an adaptive PGD-100 attack on ImageNet. The values remain far from perfect alignment, supporting the decorrelation assumption used in the analysis.}
\centering
\begin{tabular}{lc}
\toprule
Layer & Mean Correlation \\
\midrule
\texttt{conv\_proj} & 0.0897 \\
\texttt{block\_00} & 0.2107 \\
\texttt{block\_01} & 0.2518 \\
\texttt{block\_02} & 0.3311 \\
\texttt{block\_03} & 0.3712 \\
\texttt{block\_04} & 0.3781 \\
\texttt{block\_05} & 0.3083 \\
\texttt{block\_06} & 0.3484 \\
\texttt{block\_07} & 0.3683 \\
\texttt{block\_08} & 0.3784 \\
\texttt{block\_09} & 0.4056 \\
\texttt{block\_10} & 0.3670 \\
\texttt{block\_11} & 0.3624 \\
\bottomrule
\end{tabular}

\label{tab:vit_corr}
\end{table}

The mean pairwise correlations range from 0.0897 to 0.4056 across the
ViT-S layers tested. These measurements characterize alignment in this
setting. Verifying the condition above would additionally require
measuring the aggregate cross term relative to the summed squared norms.
In a separate experiment with a pre-trained DINO encoder, the $L_1$
distance between adversarial and clean embeddings decreases from 12.30
at $d=0$ to 2.12 at $d=4$ (Table~\ref{tab:sr_dino_l1}), showing that
the measured feature distortion decreases as the translation radius
increases.

\begin{table}[h]
\caption{$L_1$ distance between adversarial and clean DINO embeddings for different translation radius $d$ increases.}
\centering
\begin{tabular}{lccccc}
\toprule
Translation Radius $d$ & 0 & 1 & 2 & 3 & 4 \\
\midrule
$L_1$ Distance (Adv. vs. Clean) & 12.30 & 5.29 & 3.50 & 2.64 & 2.12 \\
\bottomrule
\end{tabular}

\label{tab:sr_dino_l1}
\end{table}

\section{Empirical Analysis for Robustness and Inference Time Trade-off}
\label{supp:trade-off}
We analyze the trade-off between the inference time and robustness incurred by our methods empirically. We note that the specific trend depends on the task, the architecture, the evaluation metric, and the attack type. Below, we present a consolidated analysis using **ResNet-50** under different translation radii \(d\), illustrating how computational budget and adversarial robustness co-vary.

\begin{table}[h]
\caption{Image Classification performance and computational cost on ImageNet with different translation radius $d$.}
\centering
\begin{tabular}{lcccc}
\toprule
\textbf{Metric} & \textbf{$d = 0$} & \textbf{$d = 1$} & \textbf{$d = 2$} & \textbf{$d = 3$} \\
\midrule
\textbf{\# Forward Passes} & 1 & 9 & 25 & 49 \\
Wall-clock Time (s) & 0.095 & 0.114 & 0.133 & 0.158 \\
SR Forward Time (excluding interpolation) & 0.007 & 0.010 & 0.019 & 0.028 \\
Top-1 Acc @ $\epsilon = 8/255$ & 8.51 & 18.40 & 20.01 & 21.17 \\
Top-1 Acc @ $\epsilon = 4/255$ & 9.45 & 19.54 & 21.02 & 22.04 \\
Top-1 Acc @ $\epsilon = 2/255$ & 11.10 & 21.02 & 22.44 & 23.44 \\
Top-1 Acc @ $\epsilon = 1/255$ & 15.21 & 24.66 & 25.80 & 26.58 \\
\bottomrule
\end{tabular}

\label{tab:sr_cost_accuracy}
\end{table}

Even at the most efficient setting $(d = 1)$, the improvement in robustness is substantial. Users can choose the desired operating point based on latency constraints and may also selectively sample transformations to further reduce computation. Notably, a significant portion of the overhead comes from translating images and performing interpolation via Python-level loops in PyTorch. The pure computation time of the network components under SR is reported as **“SR Forward Time,”** demonstrating that the method itself is efficient. Future work will focus on implementing these transformations in a custom CUDA kernel for further speedups

\section{Memory Footprint for Our Method}
\label{supp:memory}
We present the memory footprint for our method using \textbf{ResNet-50} under different translation radii $d$, evaluated on both CIFAR-sized inputs (32$\times$32) and ImageNet-sized inputs (224$\times$224). All measurements are obtained using \texttt{torch.cuda.max\_memory\_allocated(device)} to capture peak memory footprint. This illustrates how computational budget scales with SR depth.

\begin{table}[h]
\caption{Memory footprint of ResNet-50 on different size input with different degree of translation radii $d$ }
\centering
\begin{tabular}{lcccc}
\toprule
\textbf{Metric} & \textbf{$d=0$} & \textbf{$d=1$} & \textbf{$d=2$} & \textbf{$d=3$} \\
\midrule
\# Forward Passes (can be parallelized)& 1 & 9 & 25 & 49 \\
Memory (CIFAR) & 114.50 MB & 141.50 MB & 182.93 MB & 260.50 MB \\
Memory (ImageNet) & 157.80 MB & 225.19 MB & 384.03 MB & 650.47 MB \\
\bottomrule
\end{tabular}

\label{tab:memory}
\end{table}

As shown, modern GPUs have ample memory headroom, and SR’s memory growth under normal inference remains modest. The more interesting observation arises during \emph{adversarial attack generation} on one single 224$\times$224 inputs: memory usage increases dramatically, from \textbf{346.19 MB} for standard inference to \textbf{10{,}649 MB} when performing gradient-based adaptive attacks through multi-branch SR. This exponential increase makes strong adaptive attacks (“Worst-Case” in Table 1) extremely difficult in practice, highlighting an important defensive advantage of SR. 

\section{Clean Performance}
\label{supp:clean}

A potential concern with test-time defenses is that robustness may come at the expense of clean performance, particularly when the defense removes or suppresses high-frequency information. Stochastic resonance differs from smoothing-based methods in that it does not perform spatial averaging on the input image. Instead, it applies small transformations, aligns the corresponding latent embeddings, and ensembles them in feature space. As a result, it mitigates adversarial perturbations without explicitly discarding image content.

We evaluate clean performance on ImageNet using ViT-S and on CIFAR-10 using ResNet-50. As shown in Tab.~\ref{tab:clean_performance}, stochastic resonance preserves, and slightly improves, clean accuracy across both datasets. This suggests that the robustness gains are not obtained by degrading the clean signal. The strong clean-task performance observed in dense prediction experiments, including optical flow and depth estimation, further supports that high-frequency structure useful for prediction is retained.

\begin{table}[h]
\caption{\small{\bf Clean performance.} Stochastic resonance preserves and slightly improves clean accuracy on both ImageNet and CIFAR-10, indicating that the method does not trade clean performance for robustness.}
\setlength{\tabcolsep}{4pt}
\small
\centering
\begin{tabular}{l|c|c|c}
\hline
Dataset / Metric & No Defense & SR ($d=2$) & SR ($d=3$) \\
\hline
ImageNet Top-1 & 74.90 & 75.09 & \textbf{75.16} \\
ImageNet Top-5 & 92.21 & 92.30 & \textbf{92.42} \\
CIFAR-10 Accuracy & 93.26 & 93.50 & \textbf{93.64} \\
\hline
\end{tabular}

\label{tab:clean_performance}
\end{table}

\section{Additional Ablations}
\subsection{Layer-wise ablation on ResNet-50}\label{supp:layer}

We further study where stochastic resonance should be applied within a convolutional backbone. Since adversarial perturbations often behave like high-frequency artifacts, one may expect that early layers play a particularly important role: they are closest to the image lattice, preserve fine spatial structure, and determine how local perturbations are propagated into deeper semantic representations. At the same time, applying stochastic resonance too early or too late leads to different computational and representational trade-offs. Early aggregation is cheaper, while deeper aggregation acts on more semantic features but requires more repeated computation.

\begin{table}[h]
  \captionof{table}{\small{\bf Layer-wise ablation on ResNet-50.} Adversarial perturbations resemble high-frequency noise. Applying stochastic resonance through shallow layers is sufficient to defend against adversarial attacks, reducing the overall computational cost.}
\small
  \centering

  \begin{tabular}{l|c|c|c|c|c|c|c|c}
Att. Strength ($\epsilon$)&\multicolumn{2}{c|}{8/255}&\multicolumn{2}{c|}{4/255}&\multicolumn{2}{c|}{2/255}&\multicolumn{2}{c}{1/255}\\
\hline
Metric&Top-1&Top-5&Top-1&Top-5&Top-1&Top-5&Top-1&Top-5\\\hline
No Defense&8.51&26.19&9.45&27.92&11.10&30.63&15.21&36.76\\
Initial Conv.&16.91&34.66&17.36&35.28&18.12&36.30&19.59&38.25\\
Res. Block 1&22.44&44.95&23.39&46.03&24.84&47.64&27.57&50.80\\
Res. Block 2&\textbf{24.88}&\textbf{48.54}&\textbf{25.90}&\textbf{49.66}&\textbf{27.23}&\textbf{51.19}&\textbf{30.11}&\textbf{54.15}\\
Res. Block 3&21.76&44.28&22.73&45.47&24.14&47.14&27.18&50.20
  \end{tabular}

  \label{tab:layers}    
\end{table}

To examine this trade-off, we apply stochastic resonance to different termination layers of ResNet-50 on ImageNet under PGD attacks with varying perturbation strengths. As shown in Tab.~\ref{tab:layers}, applying stochastic resonance only through the initial convolution already improves robustness over the undefended model across all attack strengths. The gain becomes substantially larger when stochastic resonance is extended through the first residual block, indicating that a shallow feature ensemble is already sufficient to suppress a large fraction of the attack-induced distortion. The best performance is obtained when aggregating after the second residual block, improving Top-1 accuracy from $8.51$ to $24.88$ under $\epsilon=8/255$, and from $15.21$ to $30.11$ under $\epsilon=1/255$. The same trend holds for Top-5 accuracy.

Interestingly, extending stochastic resonance further to the third residual block does not continue to improve performance. This suggests that the benefit of stochastic resonance is not simply a function of applying it deeper in the network. Instead, there is a favorable intermediate regime where the representation is sufficiently structured to support stable alignment across translated inputs, while still remaining close enough to the spatial signal to attenuate high-frequency adversarial artifacts. Beyond this point, deeper features become increasingly semantic and less spatially equivariant, making inverse-aligned aggregation less effective.

This ablation supports two observations. First, stochastic resonance does not need to be applied through the full network to provide strong adversarial robustness. Second, the strongest setting for ResNet-50 occurs at a relatively shallow layer, namely Residual Block 2, which substantially reduces the computational overhead compared to ensembling the entire model. This makes the method practical as a test-time defense: most of the repeated computation is confined to the front end of the network, while the remaining layers are evaluated only once on the aggregated feature.

\subsection{Rotation as Purposeful Perturbations}
\label{supp:rotation}
Our main experiments use integer-pixel translations as the purposeful perturbations for stochastic resonance. This is a design choice within a broader principle: the method can be instantiated with different transformation groups, provided that the transformed features can be mapped back and aggregated in a common reference frame. In this section, we evaluate rotations as an alternative choice and explain why translations are used in the main method.

Tab.~\ref{tab:Rotation_supp} compares translations and rotations on ImageNet with ResNet-50 under PGD attacks. Rotations improve robustness over the undefended model across all attack strengths and resonance levels, showing that the stochastic-resonance mechanism is not specific to translations. However, translations provide stronger robustness at higher resonance levels and are also computationally cheaper, since rotations require interpolation. Thus, rotations constitute a valid but less effective instantiation of the same principle, rather than a failure mode of the method.

\begin{table}[h]
\caption{\small{\bf Translation vs. rotation as purposeful perturbations.} On ResNet-50, rotations improve robustness relative to the undefended model across all attack strengths and resonance levels. However, translations are more effective at higher resonance levels and avoid the interpolation overhead required by rotations. This supports our use of translations as the default instantiation of stochastic resonance.}
\setlength{\tabcolsep}{0.8pt}
\small
\centering

\begin{tabular}{l|c|c|c|c|c|c|c|c}
Att. Strength ($\epsilon$)&\multicolumn{2}{c|}{8/255}&\multicolumn{2}{c|}{4/255}&\multicolumn{2}{c|}{2/255}&\multicolumn{2}{c}{1/255}\\
\hline
Metric&Top-1&Top-5&Top-1&Top-5&Top-1&Top-5&Top-1&Top-5\\
\hline
No Defense&8.51&26.19&9.45&27.92&11.10&30.63&15.21&36.76\\
\rowcolor{gray!20}Ours ($d=1$)&18.40&40.36&19.54&42.65&21.02&43.86&24.66&47.81\\
\rowcolor{gray!20}w/ Rotation&20.01&41.32&20.84&42.39&22.10&43.89&24.72&46.89\\
Ours ($d=2$)&20.01&42.22&21.02&43.45&22.44&45.32&25.80&48.97\\
w/ Rotation&18.86&38.67&19.55&39.50&20.36&40.48&22.01&42.50\\
\rowcolor{gray!20}Ours ($d=3$)&\textbf{21.17}&\textbf{43.58}&\textbf{22.04}&\textbf{44.78}&\textbf{23.44}&\textbf{46.53}&\textbf{26.58}&\textbf{49.73}\\
\rowcolor{gray!20}w/ Rotation&15.15&32.89&15.61&33.46&16.17&34.19&17.33&35.59
\end{tabular}

\label{tab:Rotation_supp}

\end{table}

The performance gap between translations and rotations can arise from two related factors. First, convolutional backbones are naturally more compatible with translations than rotations: convolutional filters are spatially shared, but they are not inherently rotation-equivariant. Second, rotating an image on a discrete pixel grid requires interpolation, which can introduce additional artifacts before features are extracted. These two effects are difficult to disentangle completely, but both predict weaker feature alignment for rotations.
To examine this directly, we measure the alignment of transformed features after applying the inverse transformation in latent space. Specifically, for each residual block of ResNet-50, we compare the original feature with the inverse-aligned feature obtained after either translation or rotation. We report both Pearson correlation and cosine similarity in Tab.~\ref{tab:rotation_alignment}. Across all residual blocks, translated features are more strongly aligned than rotated features. This supports the interpretation that rotations provide weaker defense performance because their inverse-aligned latent features are less consistent before aggregation.

\begin{table}[h]
\caption{\small{\bf Feature alignment under translations and rotations.} We compare features after each residual block of ResNet-50, after applying the inverse transformation in latent space. Translations yield consistently higher correlation and cosine similarity than rotations, indicating better feature alignment after inversion. This explains why rotations still improve robustness relative to no defense, but are less effective than translations.}
\centering
\small
\setlength{\tabcolsep}{4pt}
\begin{tabular}{l|cc|cc}
Layer & Corr. (Trans.) & Corr. (Rot.) & CosSim (Trans.) & CosSim (Rot.) \\
\hline
Block 1 & 0.7540 & 0.6487 & 0.8357 & 0.7739 \\
Block 2 & 0.5933 & 0.4569 & 0.6756 & 0.5766 \\
Block 3 & 0.6752 & 0.5095 & 0.7771 & 0.6756 \\
Block 4 & 0.8808 & 0.7719 & 0.8954 & 0.8032 \\
\end{tabular}

\label{tab:rotation_alignment}

\end{table}

These results clarify the role of rotations in our study. The relevant comparison is not whether rotations outperform translations, but whether the stochastic-resonance principle remains meaningful under a different family of purposeful perturbations. The answer is positive: rotations improve robustness over the undefended baseline. At the same time, the lower feature-alignment scores in Tab.~\ref{tab:rotation_alignment} explain why translations are the preferred design choice for convolutional architectures. Consequently, we use translations throughout the main experiments because they instantiate the same principle more effectively, preserve latent alignment more reliably, and avoid the interpolation overhead associated with rotations.

\section{MLP-Mixer}
\label{supp:mlpmixer}

To test whether stochastic resonance depends on convolutional translation equivariance, we evaluate MLP-Mixer on CIFAR-10. We initialize the model with ImageNet pre-trained weights, fine-tune a linear head on CIFAR-10, and apply stochastic resonance at the layer before the prediction head. As shown in Tab.~\ref{tab:mlpmixer_appendix}, stochastic resonance preserves clean accuracy while substantially improving robustness under both FGSM and PGD attacks. In particular, under PGD with $\epsilon=8/255$, accuracy improves from $0.03$ to $65.38$ with $d=3$. These results suggest that while equivariance under the chosen transformations can improve feature alignment, it is not strictly required for stochastic resonance to suppress attack-induced feature distortion.

\begin{table}[h]
\caption{\small{\bf Stochastic resonance on MLP-Mixer.} Even without explicit convolutional translation equivariance, stochastic resonance substantially improves robustness under both FGSM and PGD attacks while preserving clean accuracy.}
\setlength{\tabcolsep}{3.5pt}
\small
\centering

\begin{tabular}{l|c|ccc|ccc}
\hline
\multirow{2}{*}{Setting} & \multirow{2}{*}{Clean} & \multicolumn{3}{c|}{FGSM} & \multicolumn{3}{c}{PGD} \\
& & $2/255$ & $4/255$ & $8/255$ & $2/255$ & $4/255$ & $8/255$ \\
\hline
No Defense & 89.13 & 31.41 & 24.01 & 21.76 & 0.00 & 0.01 & 0.03 \\
Ours ($d=1$) & \textbf{89.99} & 71.68 & 71.51 & 71.55 & 58.19 & 57.03 & 49.87 \\
Ours ($d=2$) & 89.94 & 73.40 & 73.30 & 73.44 & 69.08 & 68.14 & 61.41 \\
Ours ($d=3$) & 89.92 & \textbf{74.05} & \textbf{73.99} & \textbf{74.08} & \textbf{73.32} & \textbf{72.30} & \textbf{65.38} \\
\hline
\end{tabular}

\label{tab:mlpmixer_appendix}
\end{table}

\section{Implementation for Classification}
\label{supp:classification}

For our classification experiments, we built our implementation on top of standard network architectures, implementing SR on two main architectures derived from the FSR codebase \cite{kim2023feature}:
\begin{itemize}
    \item ResNet-18 \cite{he2016deep}: A standard residual network with 18 layers organized in four main blocks with increasing channel dimensions (64, 128, 256, 512).
    \item WideResNet-34 \cite{zagoruyko2016wide}: A wider variant of ResNet with depth 34 and width factor 10, resulting in higher representational capacity with three main blocks with channel dimensions scaled by the width factor (160, 320, 640).
\end{itemize}
For both architectures, we apply SR at the bottleneck layer (after the final convolutional block).

Given an input image batch $x \in \mathbb{R}^{b \times 3 \times h \times w}$ (where $b$ is the batch size), our SR approach operates as follows. First, we create a set of $(2d_x+1) \times (2d_y+1)$ perturbed versions of the input by applying pixel-level translations within the range $[-d_x, d_x] \times [-d_y, d_y]$ pixels:
\begin{equation}
    X_{\text{perturbed}} = \{g_{i,j}(X) \mid i \in [-d_y, d_y], j \in [-d_x, d_x]\}
\end{equation}
where $g_{i,j}$ translates the image by $(i,j)$ pixels. These transformations are applied using PyTorch's ``transforms.functional.affine'' function with translation parameters while preserving the original image properties.

All images are concatenated into one batch and processed through the network in parallel up to the bottleneck layer:
\begin{equation}
    F = \phi(X_{\text{perturbed}})
\end{equation}
where $\phi$ represents the network up to the bottleneck layer. This batch processing approach significantly improves computational efficiency compared to individual forward passes.

After obtaining feature maps for all perturbed inputs, we aggregate them to create a single enhanced feature map:
\begin{equation}
    F_{\text{ensembled}} = \frac{1}{n}\sum_{i,j} T_{-i,-j}(F_{i,j})
\end{equation}
where $T_{-i,-j}$ represents the inverse spatial shift operation that realigns the feature map and $n$ the number of augmentations.

Our implementation requires $(2d_x+1) \times (2d_y+1)$ forward passes through the network up to the bottleneck layer.

For evaluation, we tested our approach against standard adversarial attacks (FGSM \cite{goodfellow2014explaining}, PGD-20 and PGD-100 \cite{madry2017towards}, and C\&W \cite{carlini2017towards}), all bounded within $\epsilon = 8/255$ under $\ell_\infty$-norm. We also report an Ensemble metric that measures the worst-case performance across all attacks for each test example, providing a comprehensive robustness assessment.

\section{Implementation for Stereo Matching}
\label{supp:stereo}

For our stereo matching experiments, we built our implementation on top of standard stereo network architectures to ensure our approach remains model-agnostic and requires no training or fine-tuning. We integrated SR with PSMNet \cite{chang2018pyramid}, a pyramid stereo matching network with a stacked hourglass architecture that uses 3D convolutions on a cost volume constructed by concatenating features. We apply SR at the feature extraction stage, before cost volume construction, where stereo correspondences are first established.

Given a pair of input stereo images $x_L, x_R \in \mathbb{R}^{b \times 3 \times h \times w}$ (where $b$ is the batch size), our SR approach operates as follows. First, we create a set of $(2d_x+1) \times (2d_y+1)$ perturbed versions of each input image by applying translations within the range $[-d_x, d_x] \times [-d_y, d_y]$ pixels:
\begin{equation}
    X_{L,\text{perturbed}} = \{g_{i,j}(x_L) \mid i \in [-d_y, d_y], j \in [-d_x, d_x]\}
\end{equation}
\begin{equation}
    X_{R,\text{perturbed}} = \{g_{i,j}(x_R) \mid i \in [-d_y, d_y], j \in [-d_x, d_x]\}.
\end{equation}

All images are concatenated into batches and processed through the feature extraction component of the network:
\begin{equation}
    F_L = \phi(X_{L,\text{perturbed}})
\end{equation}
\begin{equation}
    F_R = \phi(X_{R,\text{perturbed}})
\end{equation}
where $\phi$ represents the feature extraction component of the stereo network. This batch processing approach significantly improves computational efficiency compared to individual forward passes.

After obtaining feature maps for all perturbed inputs, we aggregate them to create a single enhanced feature map.

Our implementation requires $(2d_x+1) \times (2d_y+1)$ forward passes through the feature extraction component of the network for each stereo image. 

For evaluation, we tested our approach against adversarial attacks generated using FGSM \cite{goodfellow2014explaining} and I-FGSM \cite{kurakin2018adversarial} (a special case of PGD), bounded within various $\epsilon$ values ($\{0.002, 0.005, 0.01, 0.02\}$) under $\ell_\infty$-norm. We measured performance using three standard stereo matching metrics: Mean Absolute Error (MAE), Root Mean Square Error (RMSE), and D1-error (percentage of pixels with disparity error greater than 3 pixels or 5\% of the ground truth) \cite{luo2018towards}.

\section{Details about attack on Monocular Depth Estimation}
\label{supp:mono-depth-est}
We evaluate 648 images from the NYU-V2 validation dataset. For attacks, we performed PGD attacks with $\epsilon=8/255$, $\alpha=2/255$ for 10 iterations with an L1 loss between the predicted depth and the ground truth depth. We obtain the DINOv2 features using DINOv2 backbone and perform the aggregation with these features before passing the aggregated features into the depth prediction head. We used a linear head pretrained by DINOv2 to generate the final depth prediction\footnote{\url{https://github.com/facebookresearch/dinov2\#pretrained-heads---depth-estimation}}.  

\section{Details about attack on vision language alignment}
\label{supp:vlm}
We split the COCO-Caption\footnote{\url{https://huggingface.co/datasets/lmms-lab/COCO-Caption}} validation set into the first 500 instances for evaluation and the remaining is used for generating adversarial samples for the adversarial training. For the CLIP model, we used the same CLIP model used for building the vision tower of LLaVA-13B, which is a vision transfomer with patch size 14 and input image size of 336 by 336\footnote{\url{https://huggingface.co/openai/clip-vit-large-patch14-336}}. Since CLIP aligns the feature output for the \textsc{[CLS]} token with the text embedding, we extracted spatial feature from the second last layer of the vision transformer to perform aggregation and use the last layer as it is to generate the final image embedding used for vision language alignment.  We obtain the ground truth text embedding by averaging the embedding from 5 ground truth text caption for each image provided by the dataset. For attack, we use PGD attack with $\epsilon=0.1$, $\alpha=0.02$, and 10 iterations with a cosine similarity lost between the CLIP image embedding and the ground truth text embedding.  

\section{Details about Attack on Optical Flow}
\label{supp:optical-flow}
To find an adversarial attack for optical flow estimated by a given neural network $f$, we adapt the formulation proposed by \cite{oskouie2024attack}. Specifically, we aim to find a perturbation $\delta$ for given frames $F_1$ and $F_2$, maximizing the discrepancy between predicted and ground-truth optical flow $OF$. If the ground-truth optical flow is unavailable, we use the predicted optical flow from the unattacked frame as our surrogate ground-truth. Our method applies $\delta$ to the first input frame, then uses a deep neural network to estimate optical flow from the perturbed frames. The objective is to maximize the average end-point error (EPE) between the predicted and ground-truth optical flow, calculated as the mean Euclidean distance between corresponding 2D flow vectors. In other words, the $\epsilon$-norm bounded adversary $\delta$ for optical-flow is calculated by optimizing the following
\begin{equation}
    \max_{\delta: \Vert\delta\Vert \le \epsilon} \text{EPE} \bigl(OF, f(F_1 + \delta, F_2) \bigr).  
\end{equation}

One $l_\infty$-bounded adversary $A$ for the aforementioned optimization problem is Fast Gradient Sign Method (FGSM) \cite{goodfellow2014explaining} which can be obtained by 
\begin{align}
\text{L} = \text{EPE}(OF, f(F_1, F_2)) , \nonumber \\
A = F_1 + \epsilon \cdot \text{sign}\bigl(\nabla_{F_1} \text{L} \bigr).   
\end{align}

Projected gradient descent (PGD) \cite{madry2017towards} represents an enhanced and more complex version of FGSM. This attack method generates adversarial examples through an iterative process and the formulation for this attack is as following
\begin{equation}
F_1^{(t + 1)} = \Pi_{F_1 + \mathcal{S}} \bigl( F_1^{(t)} + \alpha \cdot \text{sign}(\nabla_{F_1} \text{L} ) \bigr).   
\end{equation}

Since adversarial perturbations are constrained to be imperceptible, we assume they are too small to affect the actual physical motion between frames. Thus, we treat the ground-truth optical flow as invariant and do not update it during the optimization process.

In our experimental setup, we set the norm value to $\epsilon = \frac{10}{256}$. To generate the adversarial perturbations, we employed a 10-step PGD attack. We calculated the step size $\alpha$ as 
$2.5 \epsilon$ divided by the iteration count, ensuring a robust traversal of the $\epsilon$-ball during optimization.

\section{Feature-Space Attack and Defense}
\label{supp:feature_space}

Most adversarial attacks considered in this paper operate in pixel space and constrain the perturbation magnitude to remain visually imperceptible. Another related line of work studies attacks and defenses based on feature-space stochasticity~\cite{ukita2023adversarial}. This setting is conceptually different from ours: the feature-space attack in~\cite{ukita2023adversarial} can substantially alter the input appearance and is not constrained by the same small pixel-space perturbation assumption used in standard adversarial robustness benchmarks. Nevertheless, we include this comparison to test whether stochastic resonance can be used together with feature-space defenses and whether it remains useful when the attack mechanism differs from the pixel-space attacks used in the main paper.

We evaluate this setting on CIFAR-10. First, we apply the feature-space stochastic defense of~\cite{ukita2023adversarial} and then combine it with stochastic resonance. As shown in Tab.~\ref{tab:feature_space_joint}, stochastic resonance provides an additional improvement when used together with the feature-space defense. This indicates that our method is complementary to existing feature-space defenses, since it operates by aggregating aligned latent responses at test time rather than modifying the training objective or replacing the defense mechanism.

\begin{table}[h]
\caption{\small{\bf Combination with feature-space defense.} Stochastic resonance further improves robustness when combined with the feature-space stochastic defense of~\cite{ukita2023adversarial}, suggesting that the two mechanisms are complementary.}
\setlength{\tabcolsep}{5pt}
\small
\centering
\begin{tabular}{l|c}
\hline
Setting & Accuracy (\%) $\uparrow$ \\
\hline
Attacked & 1.72 \\
Feature-space defense~\cite{ukita2023adversarial} & 8.13 \\
Feature-space defense~\cite{ukita2023adversarial} + Ours & \textbf{8.56} \\
\hline
\end{tabular}

\label{tab:feature_space_joint}
\end{table}

We also evaluate a constrained variant of the feature-space attack, where the perturbation magnitude is bounded to better match the small-perturbation setting considered in this paper. In this setting, stochastic resonance alone achieves higher accuracy than the feature-space defense baseline, as shown in Tab.~\ref{tab:feature_space_constrained}. This result suggests that even when the attack is generated through feature-space stochasticity, the latent ensembling induced by stochastic resonance still suppresses a portion of the attack-induced feature distortion.

\begin{table}[h]
\caption{\small{\bf Constrained feature-space attack.} When the feature-space attack is constrained by a maximum perturbation magnitude, stochastic resonance alone provides a reasonable level of robustness and outperforms the feature-space defense baseline in this setting.}
\setlength{\tabcolsep}{5pt}
\small
\centering
\begin{tabular}{l|c}
\hline
Setting & Accuracy (\%) $\uparrow$ \\
\hline
Attacked & 2.14 \\
Feature-space defense~\cite{ukita2023adversarial} & 11.09 \\
Ours only & \textbf{12.36} \\
\hline
\end{tabular}

\label{tab:feature_space_constrained}
\end{table}

Overall, these results show that stochastic resonance is not restricted to the pixel-space attack mechanisms evaluated in the main paper. While our method is designed for standard imperceptible pixel-space adversarial perturbations, it can also be combined with feature-space defenses and remains effective under a constrained feature-space attack. This supports the attack-agnostic nature of stochastic resonance as a test-time defense.

\section{Boundary Handling}
\label{supp:boundary}

Our implementation uses zero filling for samples outside the image grid. To assess sensitivity to this choice, we evaluated five alternative boundary treatments under PGD-100 on CIFAR-10 using the adversarially trained ResNet-18 from Table~\ref{tab:cifar10}. Robust accuracy was 54.29\% with reflection, 54.56\% with replication, 54.55\% with circular extension, 54.40\% with valid masking, and 55.24\% with valid cropping. The five treatments differed by at most 0.95 percentage points in this setting.

\section{Additional Visualizations}
\label{supp:vis}
Here we provide additional visualizations from our experiments comparing SR under both FGSM and PGD attack. We also provide a visual showing the results of FGSM and PGD pertubation on various images.

\begin{figure*}[h]
    \centering
    \includegraphics[width=\linewidth]{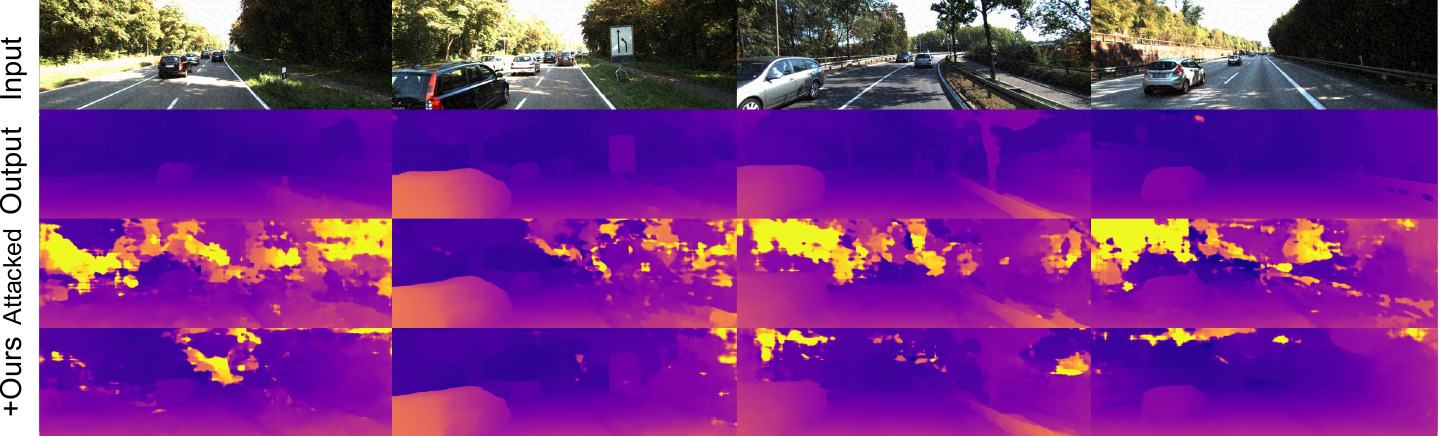}
    \caption{Visual results on stereo matching against FGSM attack, without and with SR, $\epsilon=0.02$}
    \label{fig:fgsm2e2_stereo}
\end{figure*}
\begin{figure*}[h]
    \centering
    \includegraphics[width=\linewidth]{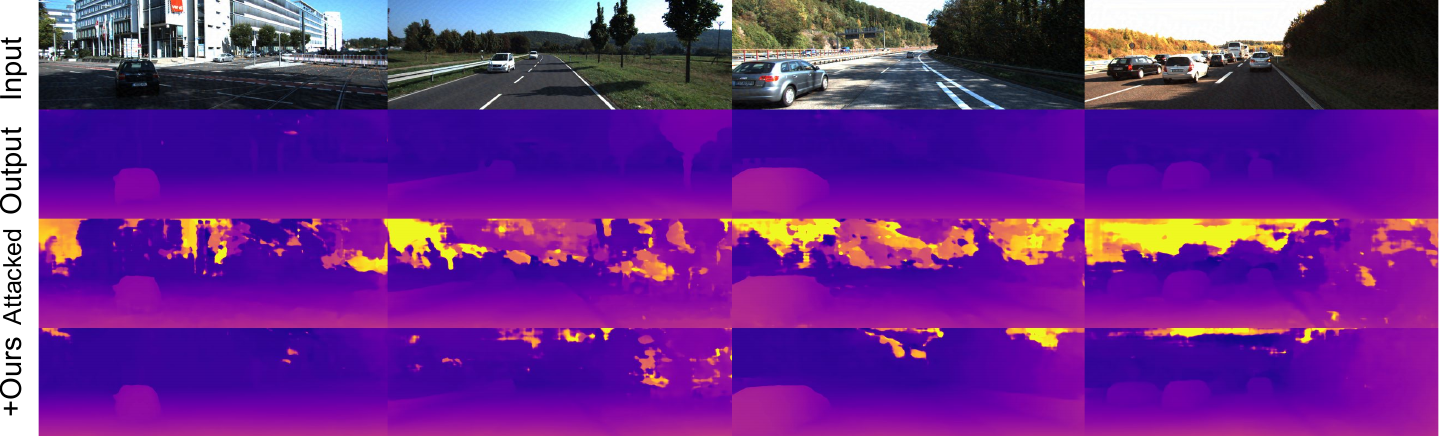}
    \caption{Visual results on stereo matching against FGSM attack, without and with SR, $\epsilon=0.01$}
    \label{fig:fgsm1e2_stereo}
\end{figure*}
\begin{figure*}[h]
    \centering
    \includegraphics[width=\linewidth]{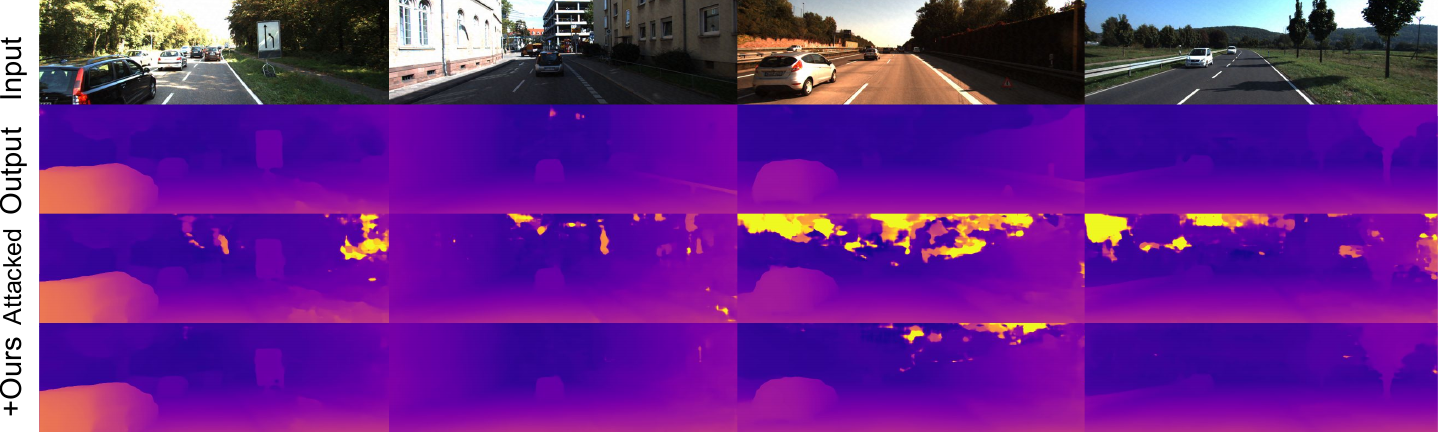}
    \caption{Visual results on stereo matching against FGSM attack, without and with SR, $\epsilon=0.005$}
    \label{fig:fgsm5e3_stereo}
\end{figure*}
\begin{figure*}[h]
    \centering
    \includegraphics[width=\linewidth]{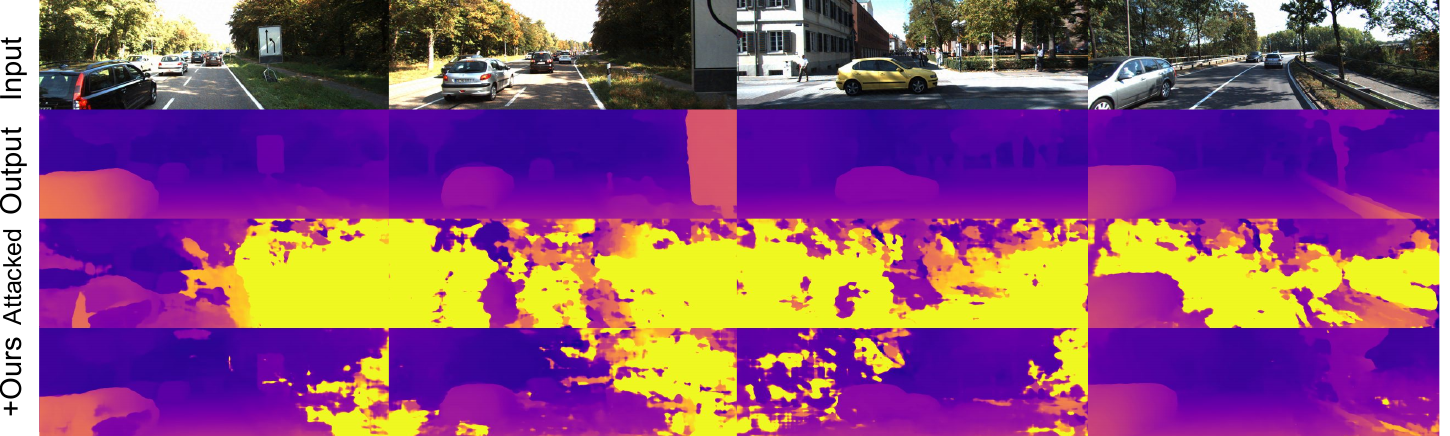}
    \caption{Visual results on stereo matching against PGD attack, without and with SR, $\epsilon=0.01$}
    \label{fig:pgd1e2_stereo}
\end{figure*}
\begin{figure*}[h]
    \centering
    \includegraphics[width=\linewidth]{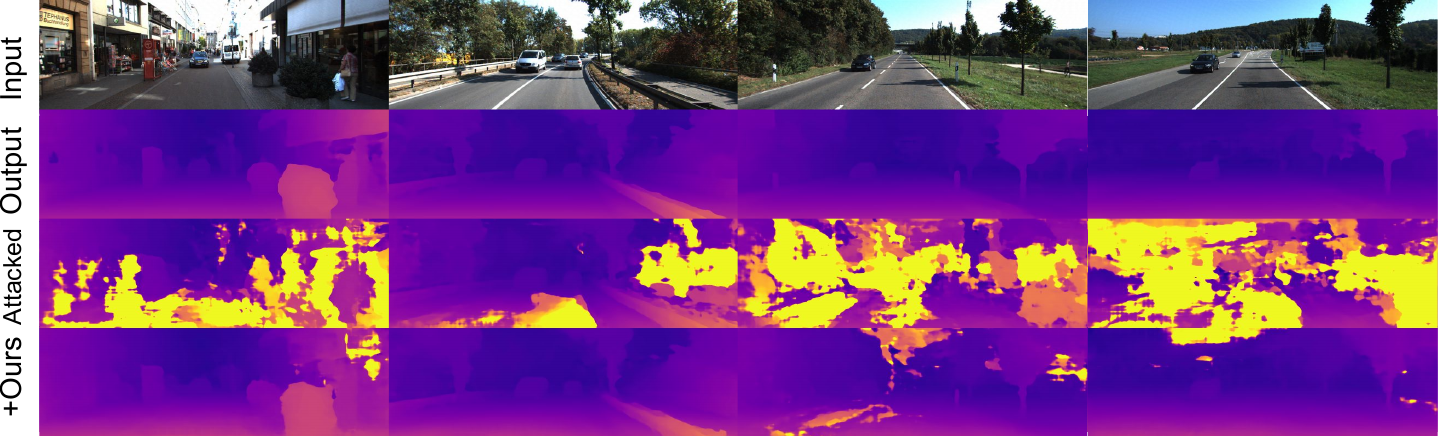}
    \caption{Visual results on stereo matching against PGD attack, without and with SR, $\epsilon=0.005$}
    \label{fig:pgd5e3_stereo}
\end{figure*}
\begin{figure*}[h]
    \centering
    \includegraphics[width=\linewidth]{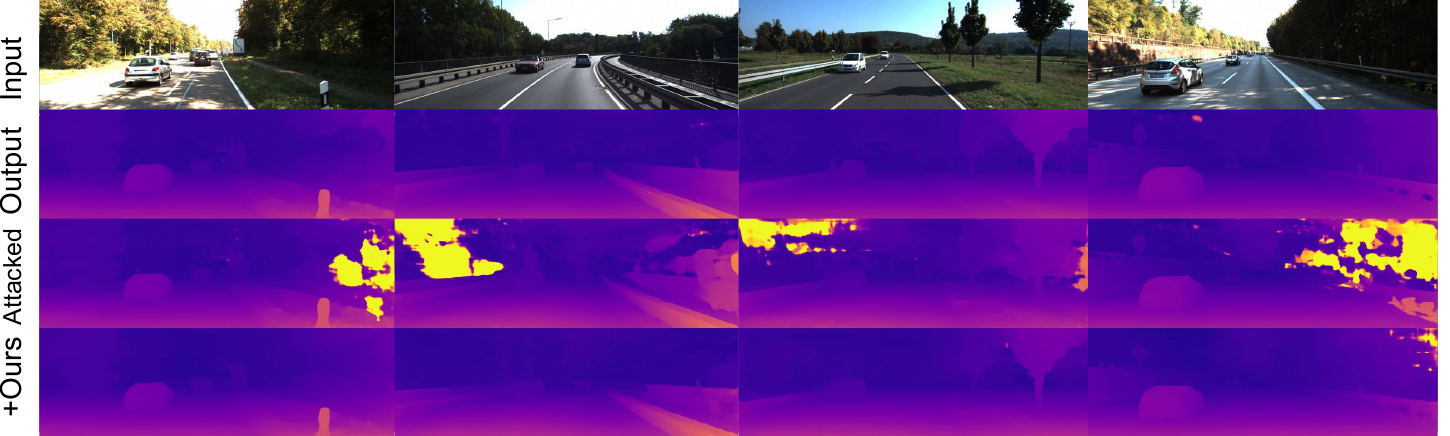}
    \caption{Visual results on stereo matching against PGD attack, without and with SR, $\epsilon=0.002$}
    \label{fig:pgd2e3_stereo}
\end{figure*}

\begin{figure}[h]
\setlength{\tabcolsep}{2pt}
    \centering
    \begin{tabular}{ccccc}
        \includegraphics[width=0.18\textwidth]{Figures/perturbations/00004.jpg} &
        \includegraphics[width=0.18\textwidth]{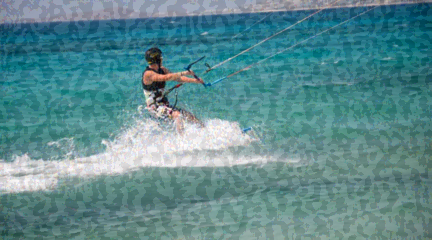} &
        \includegraphics[width=0.18\textwidth]{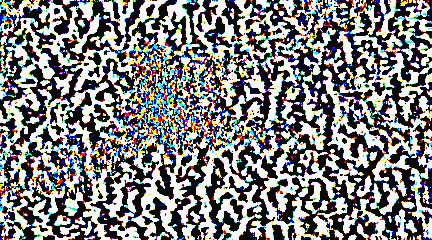} &
        \includegraphics[width=0.18\textwidth]{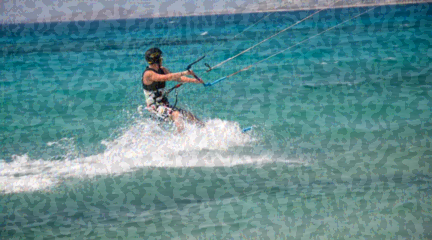} &
        \includegraphics[width=0.18\textwidth]{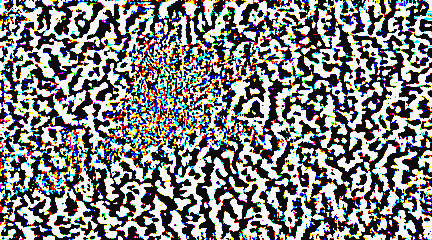} \\
        \includegraphics[width=0.18\textwidth]{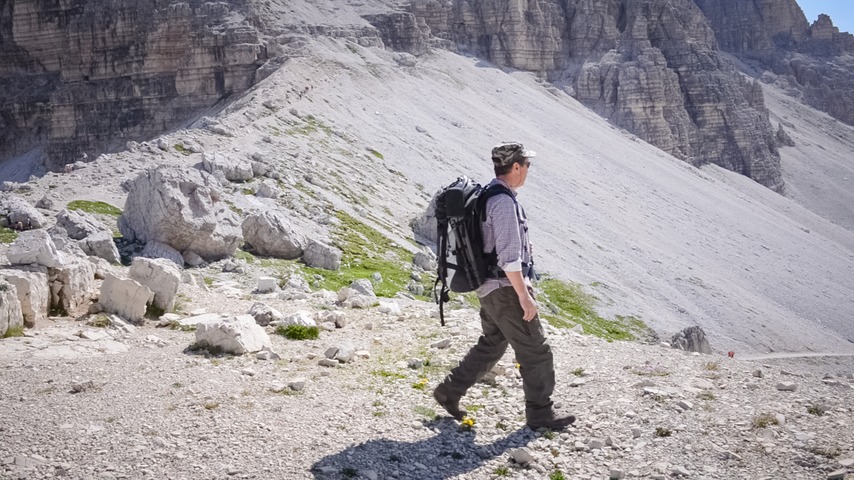} &
        \includegraphics[width=0.18\textwidth]{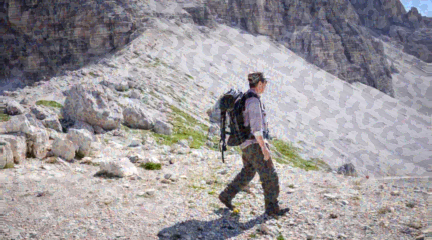} &
        \includegraphics[width=0.18\textwidth]{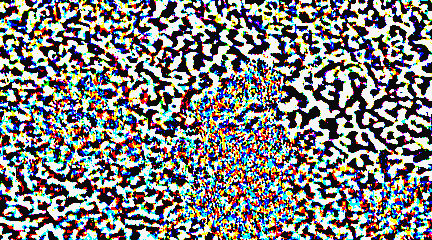} &
        \includegraphics[width=0.18\textwidth]{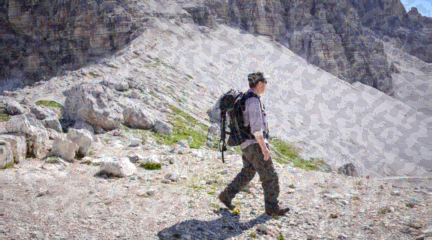} &
        \includegraphics[width=0.18\textwidth]{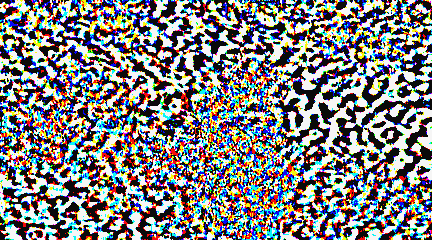} \\
        \includegraphics[width=0.18\textwidth]{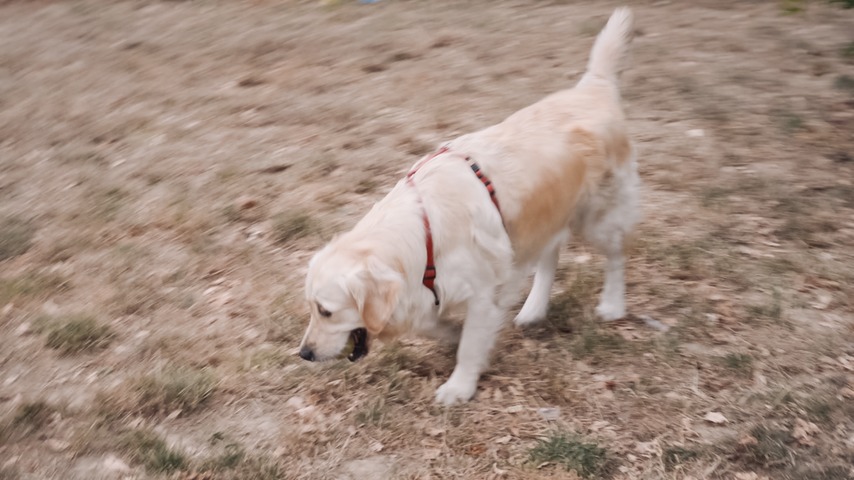} &
        \includegraphics[width=0.18\textwidth]{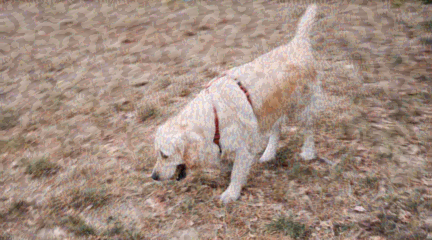} &
        \includegraphics[width=0.18\textwidth]{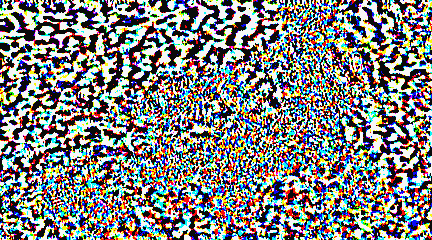} &
        \includegraphics[width=0.18\textwidth]{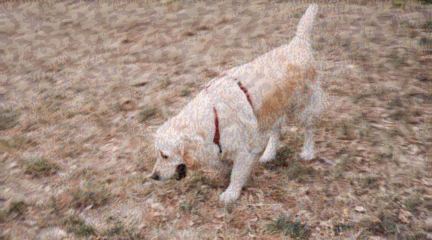} &
        \includegraphics[width=0.18\textwidth]{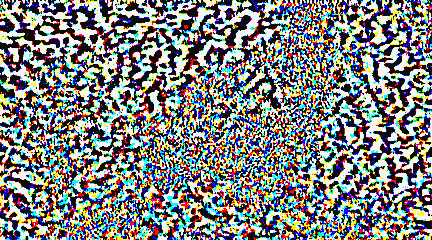} \\
        \includegraphics[width=0.18\textwidth]{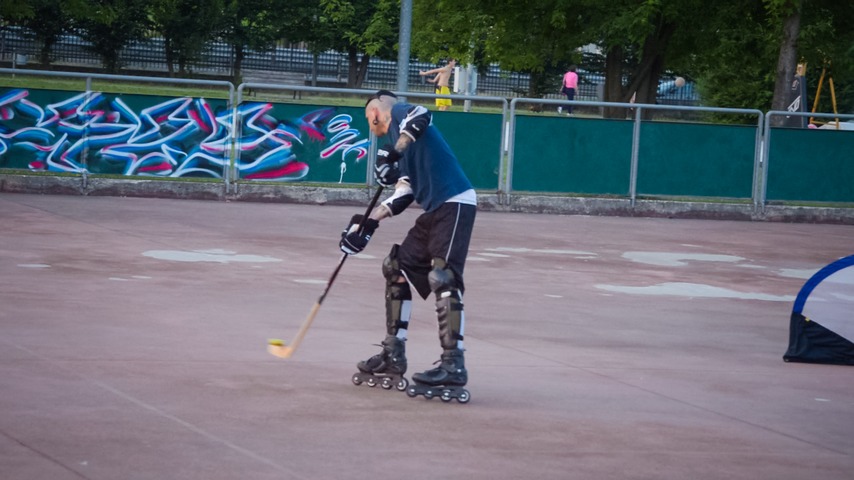} &
        \includegraphics[width=0.18\textwidth]{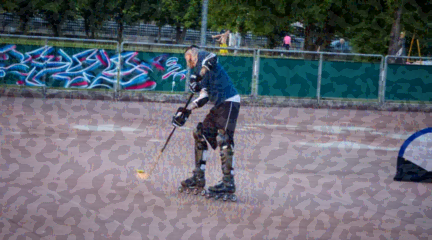} &
        \includegraphics[width=0.18\textwidth]{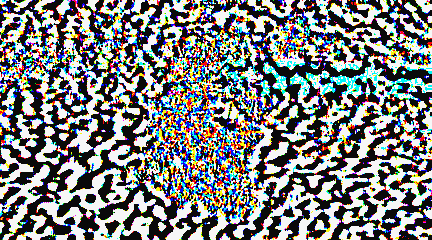} &
        \includegraphics[width=0.18\textwidth]{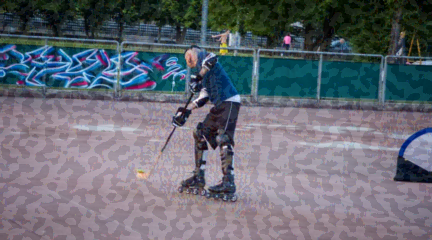} &
        \includegraphics[width=0.18\textwidth]{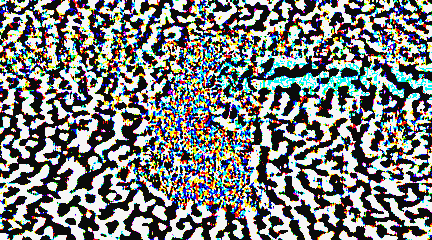} \\
        \textbf{Image} & \textbf{FGSM Attacked} & \textbf{FGSM Noise} & \textbf{PGD Attacked} & \textbf{PGD Noise} \\
    \end{tabular}
    \caption{Original images and their corresponding attacked images and perturbations using FGSM and PGD methods on optical flow. The attacks mostly target the main object observed in the image.}
    \label{Figure:attack}
\end{figure}

\end{document}